\documentclass{article} 
\usepackage[preprint]{colm2026_conference}  

\usepackage{microtype}
\usepackage{hyperref}
\usepackage{url}
\usepackage{booktabs}
\usepackage{graphicx}
\usepackage{amsmath,amssymb}
\usepackage{multirow}
\usepackage{xcolor}
\usepackage{subcaption}
\usepackage{etoc}
\usepackage{enumitem}
\usepackage{tcolorbox}
\usepackage{placeins} 

\newtcolorbox{takeawaybox}{
  colback=white,
  colframe=black,
  boxrule=0.5pt,
  left=6pt,
  right=6pt,
  top=4pt,
  bottom=4pt,
  before skip=10pt,
  after skip=10pt
}

\usepackage{lineno}

\definecolor{darkblue}{rgb}{0, 0, 0.5}
\hypersetup{colorlinks=true, citecolor=darkblue, linkcolor=darkblue, urlcolor=darkblue}

\usepackage{amsmath,amsfonts,bm}

\def\eqref#1{equation~\ref{#1}}

\def\1{\bm{1}}

\DeclareMathAlphabet{\mathsfit}{\encodingdefault}{\sfdefault}{m}{sl}
\SetMathAlphabet{\mathsfit}{bold}{\encodingdefault}{\sfdefault}{bx}{n}

\title{How to Train Your Long-Context Visual Document Model}

\author{Austin Veselka \\
LightOn \\
austin.veselka@lighton.ai
}

\begin{document}
\ifcolmsubmission
\linenumbers
\fi
\maketitle

\begin{abstract}
We present the first comprehensive, large-scale study of training long-context vision language models up to 344K context, targeting long-document visual question answering with measured transfer to long-context text.
While several such strong models are open-weight, namely Qwen3 VL and GLM 4.5/6V, their training recipes and data pipelines are not reproducible.
We systematically study continued pretraining, supervised finetuning, and preference optimization for 24B and 32B parameter models, backed by extensive LC evaluations and ablations to bridge this gap, and achieve state-of-the-art performance on MMLongBenchDoc for both parameter scales.
In addition to this, our key findings include: (i) training on context lengths that match evaluation context lengths outperforms training on longer contexts, (ii) training and evaluating with page indices provides a simple, high-impact boost to long-document performance, (iii) our synthetic data pipelines enable self-improvement via continued pretraining and supervised finetuning, and (iv) we extend the known text-to-visual long context transfer to the reverse, showing that visual long context training transfers to long-context text performance.
We also release \textsc{MMLBD-C}, a manually corrected version of MMLongBenchDoc to reduce erroneous and low quality examples in the benchmark.
\end{abstract}

\section{Introduction}
\label{sec:intro}
Long-context capabilities in large language models (LLMs) are highly desirable for applications such as summarization, in-context learning and question answering.
Thus, there is a large body of work surrounding long-context (LC) performance.
These range from cheaper attention variants \citep{linearattention,mamba} and context extension methods \citep{yarn}, to evaluations \citep{mmlbd,dude}, training data and recipes for continued pretraining (CPT) and supervised finetuning (SFT) \citep{prolong,qwen1M}, and preference optimization strategies such as LongPO \citep{longpo}.
All of these have led to significant improvements in LC performance in open models.

In enterprise and academic settings, long PDFs are common and LC vision language models (VLMs) unlock the same use cases above. Text-only LLMs struggle with these documents due to information loss and overhead from PDF to text conversion.
By comparison, VLMs are a natural fit for this use case since they process PDFs visually. However, there is a distinct lack of work on LC VLMs outside the video domain.

Until recently, closed models (e.g. GPT4o \citet{gpt4o}, Claude \citet{claude3opus}) and their newer versions have vastly outperformed open models in long-document visual question answering (VQA) on benchmarks such as MMLongBenchDoc \citep{mmlbd} and general visual LC benchmarks such as MMLongBench \citep{mmlb}.
This locked LC use cases on long PDFs to closed models.
Recently, new open-weight models, Qwen3 VL \citet{qwen3vl} and GLM 4.5/6V \citet{glm4v}, have surpassed GPT4o and achieved state-of-the-art (SOTA) performance on MMLongBenchDoc.
However, their training recipes and data strategies are underspecified and it remains unclear how to reproduce these capabilities.

To this end, this paper aims to answer the question: \emph{what works in practice for training long-context visual document models?}
Through extensive experiments on models, data and training methods with robust evaluation methodology, we produce actionable recipes for CPT, SFT, LongPO and self-improvement with SFT and quantify performance trade-offs.
Our synthetic data pipelines and a leaderboard of ablations with data compositions for each run are open sourced.\footnote{\url{https://huggingface.co/collections/lightonai/orion}, \url{https://github.com/lightonai/distilabel/tree/lc_sft_pipelines}}

\paragraph{Contributions.}
Concretely, we make the following contributions:
\begin{itemize}
  \item \textbf{Open recipes + large-scale ablations.} We provide end-to-end recipes for training long-context visual document models up to 344K context, spanning CPT, SFT, and LongPO, and report extensive ablations and compute/data trade-offs. We release the best performing Mistral and Qwen3 VL checkpoints which achieve SOTA performance for their respective model sizes on MMLongBenchDoc. We showcase our main checkpoints in Figure~\ref{fig:param_vs_mmlbdc} and Table~\ref{tab:best_ckpts}.
  \item \textbf{Page indices.} We show that adding explicit page indices is a minimal change that improves long-document VQA and long-context averages (+2.8 points on MMLBD-C and +2.8 points on visual LC average).
  \item \textbf{Targeting benchmark context length.} We show that training on context lengths suiting the benchmarks you target outperforms training on longer contexts by 1.4-3.0 points on visual LC average.
  \item \textbf{MMLBD-C.} We release \textsc{MMLBD-C}, a quality-filtered and corrected evaluation variant of MMLongBenchDoc, modifying 251 out of 1091 examples for errors, incorrect grammar or misleading/underspecified questions and removing 16.
  \item \textbf{Self-improvement.} We demonstrate that our synthetic data pipelines enable self-improvement, with CPT improving over Mistral Small 3.1 Instruct\footnote{\url{https://huggingface.co/mistralai/Mistral-Small-3.1-24B-Instruct-2503}}, hereon 'Mistral', by +3.8 points and SFT improving by +3.2 points on visual LC average.
  \item \textbf{Visual LC to Text LC Transfer.} We demonstrate that long-document VQA training transfers strongly to long-context text performance (+11.5 points on Helmet \citet{helmet}), the reverse of text to vision transfer shown in \citep{text_to_visual_lc}.
\end{itemize}

\section{Related work}
\label{sec:related}
We now situate our contributions within the broader literature on long-context modeling, synthetic data, and evaluation.


\paragraph{Long-context VLMs and long-document understanding.}
Much of existing work on LC VLMs has focused on long video processing \citep{longvila,longvita,bolt,temporalcot,infinitevl,vocollama}. In contrast, long-document understanding has obvious applications in enterprise and academic settings, but has remained mainly the strength of closed or open-weight models \citep{gpt41,qwen3vl,glm4v}, though some recent work has explored this setting \citep{docopilot,v2pe}. Docopilot is particularly similar, they introduce a large dataset of long image documents from ArXiv, Sci-Hub and OpenReview and fine tune InternVL 2 2B. We surpass this work in scale.

\paragraph{Synthetic data.}
Synthetic data has become the primary mechanism for scaling instruction tuning since human labeled data is expensive and time-consuming to collect. Early techniques focus on instruction, or question, generation \citep{selfinstruct,evolinstruct,magpie}. Various other works apply techniques specifically designed for long-context data generation \citep{longalign,nextlong,wildlong,bootstrap}. We propose new pipelines for challenging multi-page question generation and recursive answer generation allowing for weak to strong self-improvement.


\paragraph{Self-improvement for long-context.}
Long context in particular is well suited for self-improvement: since models still suffer strongly from decaying performance with increasing context length, there is room for improvement simply by generalizing short-context capabilities to long-context. Recently, \citep{longpo, solopo} have explored preference optimization with methods applicable to this setting, showing strong results and outperforming SFT on LC text benchmarks. We apply LongPO to long-document VQA at a large scale, high context length, and compare it to SFT.

\paragraph{Evaluation for long-context vision and text.}
Evaluation for LC initially focused on toy needle-in-a-haystack (NIAH) tasks \citep{og_niah,ruler}, however NIAH is easily saturated, so LC benchmarks have evolved to include more challenging and realistic tasks \citep{helmet,longbench_v2,infinitybench}. Evaluation for LC VLMs outside of video benchmarks has improved significantly, with DUDE and more recently, MMLongBenchDoc \citet{mmlbd} and MMLongBench \citet{mmlb}.


\section{Background and setup}
\label{subsec:setup}
\begin{figure}[t]
\begin{center}
\includegraphics[width=0.85\linewidth]{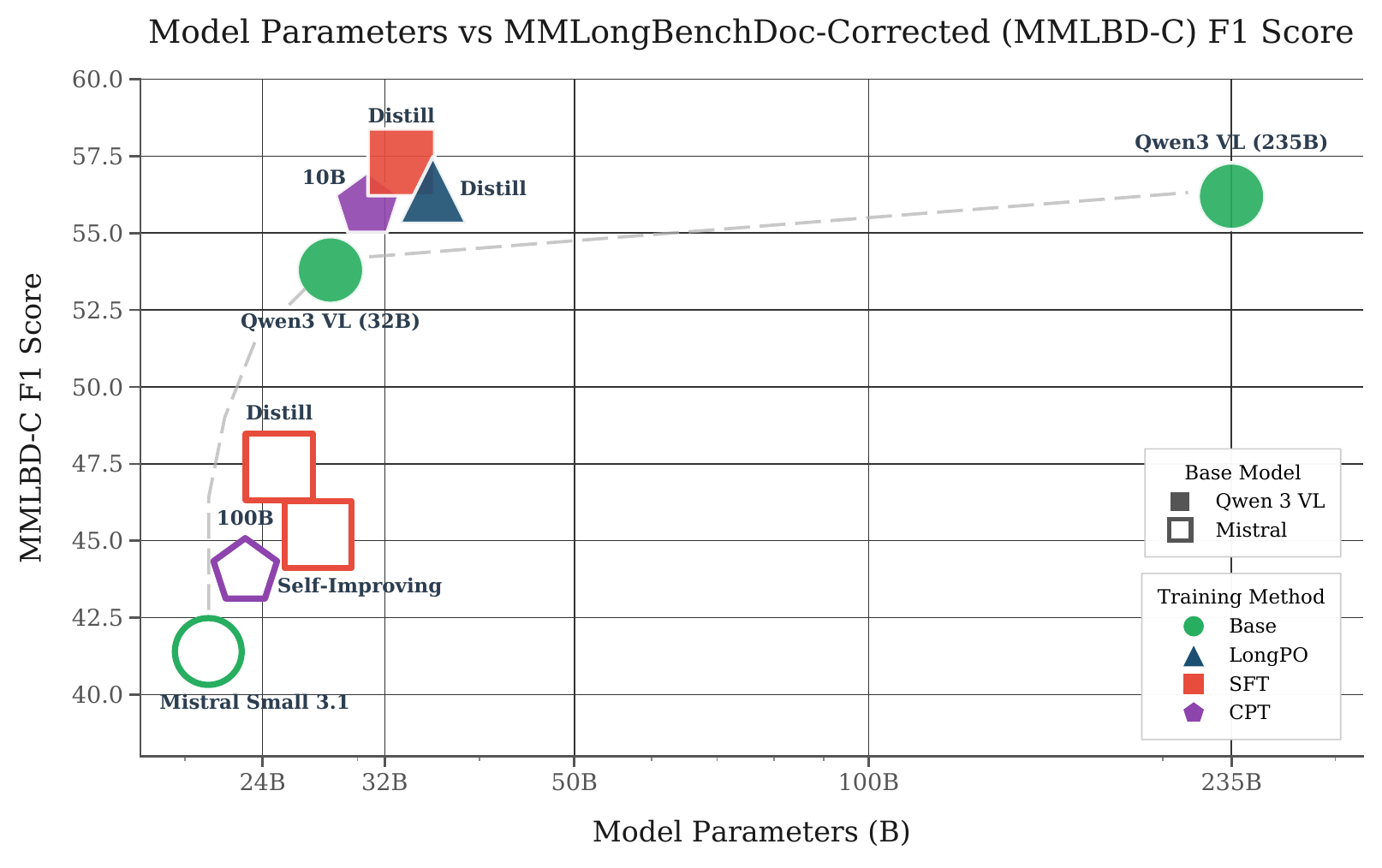}
\end{center}
\caption{Performance for our best training recipes compared to the base models we train and with the previous SOTA Qwen3 VL 235B A22B. We set a new SOTA on this version of MMLongBenchDoc \citet{mmlbd} with SFT + CPT outperforming LongPO. 'Distill' describes the answer generation pipeline. We include scores for the self-improving setting using Mistral and its CPT checkpoint for answer generation with our recursive pipeline. See Appendix~\ref{pa:main_recipes} for specific training recipes.}
\label{fig:param_vs_mmlbdc}
\end{figure}

We now describe the key components of our experimental setup: data collection and training methodology along with the necessary background for the preference optimization method we use, LongPO.

\paragraph{Data.}
\label{pa:data}
Following the work of ColPali \citet{colpali}, we construct a foundational corpus of PDFs for training by generating detailed search queries, scraping the web and filtering. We end up with a corpus of 250K PDFs and 16M pages, which serves as a foundation for synthetic LC examples from real world long documents. We augment this with the PDFA English split \citet{pdfa}, which contains 2M PDFs and 18M pages.
Additional details on the construction and makeup of the corpus are provided in Appendix~\ref{sec:corpus_details}.

\paragraph{Multi-stage training.}
Training at high sequence lengths, e.g. 344K tokens for Mistral, requires high degrees of sequence parallelism (SP). To mitigate communication overhead from this, we split all training into two stages: a short stage with examples of up to 104 pages and a long stage with examples of up to 336 pages. See Appendix~\ref{pa:training_details} for page resolution details.

\paragraph{Model merging.}
\label{pa:model_merging}
For all models and training methods (CPT, SFT, LongPO), we find that training results in catastrophic forgetting and degrades model performance. However, when we apply model merging \citep{task_arithmetic}, we find that we can improve results without degrading the normal instruct performance of the model. See Appendix~\ref{sec:repro} for more details.

\paragraph{LongPO.}
\label{pa:longpo}
LongPO \citep{longpo} is a preference optimization method based on DPO \citet{dpo}, adapted to extend short context performance to LC inputs. Briefly, LongPO generates chosen and rejected responses from short-context, used to generate the instruction, and long-context inputs respectively. To counter out of distribution scores from the reference model which is less adept at long context, LongPO derives the training objective with a \textit{short-to-long constraint}: effectively, the reference model scores are derived from the short context, rather than the long context. The LongPO objective is:
\begin{equation}
\label{eq:longpo_objective}
\resizebox{0.9\linewidth}{!}{$\displaystyle
\mathcal{L}_{\text{LongPO}} = - \lambda \mathbb{E} \left[ \log \sigma \left( \beta \log \frac{\pi_\theta(y_w \mid x_L)}{\pi_{\text{ref}}(y_w \mid x_S)} - \beta \log \frac{\pi_\theta(y_l \mid x_L)}{\pi_{\text{ref}}(y_l \mid x_S)} \right) \right] + \mathcal{L}_{\text{NLL}}
$}
\end{equation}
where $y_w, y_l$ are chosen/rejected responses from short/long context inputs $x_S, x_L$.
They weight the preference objective with $\lambda = 0.01$.

\section{Evaluation protocol}
\label{sec:eval}
Conducting large sets of ablations requires diverse benchmarks to reduce noise and avoid overfitting, so we employ a suite of long-context benchmarks that target visual and textual LC performance while focusing on long document understanding and propose the following two aggregates as metrics:
\begin{description}[nosep,leftmargin=1em,labelindent=0em]
    \item[Visual-LC Avg (VA)]: averaged across MMLongBenchDoc \citet{mmlbd}, MMLBD-C, MMLongBench \citet{mmlb}, DUDE \citet{dude} and SlideVQA \citet{slidevqa}.
    \item[LC Avg (LCA)]: visual-LC benchmarks, HELMET \citet{helmet} and LongBench v2 \citet{longbench_v2}.
\end{description}

Since these benchmarks have different score distributions, we normalize the scores by the maximum score for each benchmark before averaging to ensure a balanced comparison. Qwen3 VL 235B A22B is typically the upper bound for each benchmark, ensuring these aggregates are stable under new experiments. Since we focus on long-document VQA, VA will be our primary metric, with MMLBD-C as tiebreaker since this is the most relevant benchmark to our work. Across 3 runs, our metrics are stable: VA has $\sigma = 0.33$ and LCA has $\sigma = 0.24$. See Appendix~\ref{sec:evaluation_benchmarks} for a full list of benchmarks and details. We note that MMLBD-C scores correlate highly with MMLongBenchDoc scores, though MMLBD-C scores are generally higher.

\subsection{MMLBD-C: correcting MMLongBenchDoc}
\label{sec:mmlbdc}
We construct \textsc{MMLBD-C} by flagging and correcting issues in MMLongBenchDoc including incorrect question-document pairing, ambiguous or misleading wording, typos, and answer errors. To do this, we apply a version of the recursive pipeline (see Section~\ref{sec:answer_generation} and Appendix~\ref{sec:appendix_alternative_synthetic_data_pipelines}) adapted to find inconsistencies between the source, question and answer. A total of 342 examples are flagged, which we manually review and take one of the following actions: leave as is, modify the question or answer, or remove from the benchmark. In total, we modify 251 examples and remove 16. We include examples below and release the annotations for public inspection. See images in Appendix Figure~\ref{fig:mmlbdc_examples}.

\begin{description}
  \item[\textit{Document mismatch}:] ``List all the PM health effects that increse by more than 35\% in India and Thailand.'' was paired with an unrelated document about digital marketing. We remove 9 of 10 affected questions and convert the last to 'Not answerable'.
  \item[\textit{Underspecified}:] ``List all the sections that discuss about the experiment setup?''. Answer: ``['Section 4.1', 'Section 4.2', 'Section 4.3', 'Appendix A']''. It is hard to argue the Methodology section does not discuss the experiment setup.
  \item[\textit{Typo}:] ``How do Amazon recognize \underline{least} cost?'' should read ``\underline{lease} cost''; since least is a reasonable word in this context, the model can be justifiably confused.
  \item[\textit{Incorrect answer}:] ``How many percentage respondents in this survey access to internet more than two times per month?'' was marked unanswerable despite explicit evidence in the document.
  \item[\textit{Answer expansion}:] For ``Not answerable'' questions, we also accept equivalent responses, e.g. ``None'', ``0'' or ``No one'', where appropriate.
\end{description}

\section{Long document VQA training approaches}
\label{sec:long_document_vqa_training_approaches}
Throughout this section we discuss the four training settings we explore: CPT, SFT, LongPO and self-improvement with SFT. Our goal is to measure the performance and compute of each method, discover practical techniques and recipes, evaluate the weak-to-strong LC capabilities of our synthetic data pipelines with self-improvement and produce strong models according to our VA metric.

\subsection{Continued pretraining (CPT)}
\label{sec:cpt}
We begin with CPT to extend the context length of Mistral and investigate the impact of CPT on visual and text LC performance, we adopt LC text data from Prolong \citet{prolong}, adapt the tasks from Qwen-2.5-1M \citet{qwen1M} to the visual domain and introduce a novel task for counting. These tasks are:

\begin{itemize}
  \item \textbf{Fill-in-the-middle (FIM)}: we remove a page, parse the content with Mistral and train the model to fill in the missing text.
  \item \textbf{Unshuffle}: a visual version of paragraph re-ordering, where the model must predict the correct order for a shuffled document.
  \item \textbf{Key/position-based retrieval}: a visual version of key/position-based retrieval where the model must retrieve text near a given key or described by a certain position.
  \item \textbf{Counting}: a novel task where a model labels the count of an instance on each page individually, then a LC example is constructed with a chain of thought \citet{cot} that lists the count for each page and the final sum.
\end{itemize}

Excepting counting, these tasks produce extremely scalable pretraining datasets, due to requiring annotation of only a single page per long-context example, or being entirely programmatic in the case of unshuffle. From this data, we study (i) the minimal necessary CPT to achieve strong performance, (ii) the impact of each task (drop-one ablations), and (iii) visual LC to text LC transfer. Length distributions for CPT and SFT examples are provided in Appendix~\ref{sec:appendix} and we include an ablation on the transferability of CPT to different model families using Qwen3 VL 32B Instruct\footnote{\url{https://huggingface.co/Qwen/Qwen3-VL-32B-Instruct}}, hereon 'Qwen3 VL', in Appendix \ref{sec:cpt_qwen3vl}.

\subsubsection{Minimal CPT scale}
\label{sec:minimal_cpt_scale}
We CPT Mistral Base at total token horizons, image + text, of 1B, 10B and 100B. The results are shown in Table \ref{tab:cpt_scale}. We find that training on 1B tokens achieves a similar VA score to the 10B checkpoint, however, the 1B checkpoint underperforms on MMLBD-C which is an important target for our work. Compared to the 1B checkpoint, 10B and 100B tokens continue to deliver gains, with LCA scores increasing smoothly with scale and VA improving at 100B tokens.

We additionally investigate the impact of skipping CPT entirely. We train two checkpoints on the same data, starting from Mistral Instruct and from the merged 100B CPT checkpoint and show results in Table \ref{tab:no_cpt}. Surprisingly, we find that SFT alone is competitive with SFT and CPT, with the main exception being HELMET. This shows that SFT and CPT are not additive in our setting for most benchmarks. We provide additional analysis on extended context lengths when using CPT vs SFT alone in Appendix \ref{sec:extended_context_analysis}.

\begin{takeawaybox}
  \textbf{Takeaway \#1:} If compute is constrained or the model's context length is sufficient, CPT may be skipped---SFT alone achieves competitive results in the targeted area (VA). However, CPT data is extremely scalable, requires no strong teacher and improves LC text performance: $+4.9-7.3$ points on HELMET.

\end{takeawaybox}

\subsubsection{Impact of each CPT task}
\label{sec:cpt_tasks}
We perform a series of ablations on the impact of each CPT task by removing one task at a time from the 10B set and CPT on the rest. We find the following ranking of task importance based on the VA score (most to least impactful; see Table~\ref{tab:cpt_ablations} in Appendix):
\begin{center}
Fill in Middle $\succ$ Unshuffle $\succ$ Key/Position Retrieval $\succ$ Prolong LC Text $\succ$ Sum Count
\end{center}
\vspace{-0.5em}
though we note that each of these performs worse than the combined set.

\subsubsection{Visual LC to text LC transfer}
\label{sec:visual_to_text_transfer}
Motivated by the findings of \citep{text_to_visual_lc} that training on LC text data extends the context length on video data, we ask the reverse question: \emph{does training on visual LC only improve LC text performance?}. We apply CPT to Mistral without prolong text data for 10B tokens and measure an increase in HELMET scores from 37 to 48.5, showing that visual LC training benefits LC text understanding. This also shows that the large improvements in HELMET due to CPT are not entirely the result of the included LC data.

\begin{table}[t]
  \begin{center}
  \resizebox{\textwidth}{!}{%
  \begin{tabular}{lcccccccc}
  \toprule
  \multicolumn{1}{l}{\bf Checkpoint} & \multicolumn{1}{c}{\bf VA} & \multicolumn{1}{c}{\bf LCA} & \multicolumn{1}{c}{\bf MMLBD-C} & \multicolumn{1}{c}{\bf MMLB 128K} & \multicolumn{1}{c}{\bf SlideVQA} & \multicolumn{1}{c}{\bf Helmet} & \multicolumn{1}{c}{\bf LongBench v2} & \multicolumn{1}{c}{\bf DUDE} \\
  \midrule
  Qwen3 VL (10B) & 92.9 \textcolor{red}{(-1.3)} & 91.6 \textcolor{red}{(-1.1)} & 55.9 \textcolor{teal}{(+2.1)} & 72.0 \textcolor{teal}{(+1.6)} & 69.7 \textcolor{red}{(-7.5)} & 64.6 \textcolor{teal}{(+1.7)} & 40.0 \textcolor{red}{(-2.0)} & 57.5 \textcolor{red}{(-4.2)} \\
  Mistral (100B) & 84.4 \textcolor{teal}{(+3.8)} & 84.6 \textcolor{teal}{(+7.5)} & 42.7 \textcolor{teal}{(+1.3)} & 69.3 \textcolor{teal}{(+2.9)} & 68.0 \textcolor{teal}{(+0.2)} & 51.7 \textcolor{teal}{(+14.6)} & 47.0 \textcolor{teal}{(+8.0)} & 60.1 \textcolor{teal}{(+7.3)} \\
  Mistral (1B) & 83.5 \textcolor{teal}{(+2.9)} & 80.1 \textcolor{teal}{(+3.0)} & 40.8 \textcolor{red}{(-0.6)} & 70.1 \textcolor{teal}{(+3.7)} & 70.6 \textcolor{teal}{(+2.8)} & 40.4 \textcolor{teal}{(+3.4)} & 40.0 \textcolor{teal}{(+1.0)} & 57.4 \textcolor{teal}{(+4.6)} \\
  Mistral (10B) & 83.4 \textcolor{teal}{(+2.8)} & 81.3 \textcolor{teal}{(+4.2)} & 43.1 \textcolor{teal}{(+1.7)} & 68.1 \textcolor{teal}{(+1.7)} & 66.9 \textcolor{red}{(-0.9)} & 46.0 \textcolor{teal}{(+8.9)} & 41.0 \textcolor{teal}{(+2.0)} & 55.1 \textcolor{teal}{(+2.3)} \\
  \bottomrule
  \end{tabular}
  }%
  \end{center}
  \caption{CPT at different token horizons with deltas between the checkpoint and the base model (Mistral Base or Qwen3 VL Instruct).}\label{tab:cpt_scale}
  \end{table}

\begin{table}[t]
  \begin{center}
  \resizebox{\textwidth}{!}{%
  \begin{tabular}{lcccccccc}
  \toprule
  \multicolumn{1}{l}{\bf Checkpoint} & \multicolumn{1}{c}{\bf VA} & \multicolumn{1}{c}{\bf LCA} & \multicolumn{1}{c}{\bf MMLBD-C} & \multicolumn{1}{c}{\bf MMLB 128K} & \multicolumn{1}{c}{\bf SlideVQA} & \multicolumn{1}{c}{\bf Helmet} & \multicolumn{1}{c}{\bf LongBench v2} & \multicolumn{1}{c}{\bf DUDE} \\
  \midrule
  SFT Only & 84.8 \textcolor{teal}{(+1.9)} & 81.6 \textcolor{teal}{(+0.1)} & 45.4 \textcolor{teal}{(+0.7)} & 71.7 \textcolor{teal}{(+1.2)} & 70.9 \textcolor{teal}{(+9.1)} & 47.1 \textcolor{red}{(-7.3)} & 37.0 & 53.5 \textcolor{red}{(-1.5)} \\
  SFT from CPT & 84.4 \textcolor{teal}{(+1.5)} & 82.2 \textcolor{teal}{(+0.7)} & 45.1 \textcolor{teal}{(+0.4)} & 67.9 \textcolor{red}{(-2.5)} & 72.9 \textcolor{teal}{(+11.2)} & 52.0 \textcolor{red}{(-2.4)} & 37.0 & 53.9 \textcolor{red}{(-1.0)} \\
  SFT from Instruct + CPT & 82.9 & 81.5 & 44.6 & 70.4 & 61.7 & 54.4 & 37.0 & 54.9 \\
  \bottomrule
  \end{tabular}
  }%
  \end{center}
  \caption{Comparison of SFT performance with and without CPT.}\label{tab:no_cpt}
\end{table}

\subsection{Supervised finetuning (SFT)}
\label{sec:sft}
Having established CPT's role in extending context length and improving LC text performance, we now turn to SFT, aiming to find the most effective synthetic data pipelines for visual LC. We break this down into question generation methods with Magpie \citet{magpie} as a baseline, and answer generation methods with distillation from a strong teacher model, here Qwen3 VL 235B A22B \citet{qwen3vl}, as a baseline.

We found question generation to have a minor impact on VA performance, so we defer the details to Appendix~\ref{sec:question_generation_details}. We also study the impact of training context length, page indices during training and evaluation and the base model for training. Additional experiments can be found in Appendix \ref{sec:sft_experiments}.


\subsubsection{Answer generation}
\label{sec:answer_generation}
For answer generation we employ one of two pipelines: the first is a recursive pipeline which extracts evidence relevant to the given question from each page individually, uses a numerical score from the extraction model to rank the pages by relevance and passes either the most relevant pages or their extracted evidence to Qwen3 VL 235B A22B or Qwen3 235B respectively. As a baseline, the second method passes the full example to Qwen3 VL 235B A22B, which we refer to as plain distillation.

To compare these, we train Qwen3 VL on 50K samples from each pipeline. As shown in Table~\ref{tab:recursive_vs_plain_distill}, the recursive pipeline outperforms in VA and LCA, specifically in MMLongBench, SlideVQA and LongBench v2, while underperforming on MMLBD-C. We include an ablation on the same experiment with LongPO in Appendix \ref{sec:longpo_recursive_vs_plain_distill}. In addition, we will later show that the recursive pipeline enables self-improvement on VA and LCA.

\begin{table}[t]
  \begin{center}
  \resizebox{\textwidth}{!}{%
  \begin{tabular}{lcccccccc}
  \toprule
  \multicolumn{1}{l}{\bf Checkpoint} & \multicolumn{1}{c}{\bf VA} & \multicolumn{1}{c}{\bf LCA} & \multicolumn{1}{c}{\bf MMLBD-C} & \multicolumn{1}{c}{\bf MMLB 128K} & \multicolumn{1}{c}{\bf SlideVQA} & \multicolumn{1}{c}{\bf Helmet} & \multicolumn{1}{c}{\bf LongBench v2} & \multicolumn{1}{c}{\bf DUDE} \\
  \midrule
  Recursive & 92.2 \textcolor{teal}{(+1.1)} & 92.6 \textcolor{teal}{(+1.9)} & 54.5 \textcolor{red}{(-2.5)} & 72.0 \textcolor{teal}{(+4.2)} & 74.2 \textcolor{teal}{(+8.6)} & 65.9 \textcolor{teal}{(+0.5)} & 45.0 \textcolor{teal}{(+4.0)} & 55.3 \textcolor{red}{(-0.7)} \\
  Plain Distillation & 91.1 & 90.7 & 57.0 & 67.8 & 65.6 & 65.4 & 41.0 & 55.9 \\
  \bottomrule
  \end{tabular}
  }%
  \end{center}
  \caption{Comparison of answer generation pipelines: recursive vs plain distillation.}\label{tab:recursive_vs_plain_distill}
\end{table}

\subsubsection{Targeting benchmark context length}
\label{sec:targeting_benchmark_context_length}
Previous work shows that training on contexts longer than evaluation is beneficial for performance \citep{prolong}.
However, we find from multiple experiments that training on context lengths similar to the benchmarks outperforms training on longer contexts. With SFT on Mistral, we find that training on only the short stage (up to 104 pages) is stronger than training on both stages (up to 336 pages). As shown in Table \ref{tab:short_vs_long_context}, the short stage only model improves scores across the board.
We measure this result in two additional scenarios: SFT and LongPO on Qwen3 VL and we see the same trend.

\begin{table}[t]
  \begin{center}
  \resizebox{\textwidth}{!}{%
  \begin{tabular}{lcccccccc}
  \toprule
  \multicolumn{1}{l}{\bf Checkpoint} & \multicolumn{1}{c}{\bf VA} & \multicolumn{1}{c}{\bf LCA} & \multicolumn{1}{c}{\bf MMLBD-C} & \multicolumn{1}{c}{\bf MMLB 128K} & \multicolumn{1}{c}{\bf SlideVQA} & \multicolumn{1}{c}{\bf Helmet} & \multicolumn{1}{c}{\bf LongBench v2} & \multicolumn{1}{c}{\bf DUDE} \\
  \midrule
  Mistral Short Stage & 84.6 \textcolor{teal}{(+3.0)} & 83.8 \textcolor{teal}{(+3.5)} & 45.0 \textcolor{teal}{(+1.7)} & 67.1 \textcolor{teal}{(+2.4)} & 72.0 \textcolor{teal}{(+6.1)} & 54.6 \textcolor{teal}{(+4.0)} & 41.0 \textcolor{teal}{(+2.0)} & 55.4 \textcolor{teal}{(+1.0)} \\
  Mistral Short Stage + Long Stage & 81.5 & 80.2 & 43.3 & 64.7 & 65.9 & 50.6 & 39.0 & 54.4 \\
  \midrule
  Qwen3VL Short Stage & 92.5 \textcolor{teal}{(+1.4)} & 92.5 \textcolor{teal}{(+1.9)} & \textbf{57.3} \textcolor{teal}{(+0.3)} & 73.8 \textcolor{teal}{(+6.0)} & 66.8 \textcolor{teal}{(+1.3)} & 65.7 \textcolor{teal}{(+0.3)} & 44.0 \textcolor{teal}{(+3.0)} & 54.8 \textcolor{red}{(-1.2)} \\
  Qwen3VL Short Stage + Long Stage & 91.1 & 90.7 & 57.0 & 67.8 & 65.6 & 65.4 & 41.0 & 55.9 \\
  \midrule
  LongPO Short Stage & 94.6 \textcolor{teal}{(+2.2)} & 93.1 \textcolor{teal}{(+1.6)} & 56.4 \textcolor{teal}{(+2.4)} & 75.6 \textcolor{teal}{(+3.8)} & 75.5 \textcolor{teal}{(+0.7)} & 62.9 \textcolor{red}{(-0.5)} & 42.0 & 56.0 \textcolor{teal}{(+1.9)} \\
  LongPO Short Stage + Long Stage & 92.4 & 91.5 & 54.0 & 71.8 & 74.8 & 63.4 & 42.0 & 54.1 \\
  \bottomrule
  \end{tabular}
  }%
  \end{center}
  \caption{Training on short stage only (up to 104 pages) vs both stages (up to 336 pages) for SFT on Mistral, Qwen3 VL and LongPO on Qwen3 VL.}\label{tab:short_vs_long_context}
\end{table}

We reconcile this apparent contradiction by examining the training data distributions. While ProLong's 512K stage has a maximum sequence length of 512K tokens, the mean and median are only 1,262 and 484 tokens respectively---the distribution is heavily short-skewed, with the vast majority of examples being short. In contrast, our long stage contains genuinely long examples with a median of 156 images per example, note the benchmarks are mostly under 128K tokens. We include a table comparing the training data distributions in Appendix Table~\ref{tab:prolong_comparison}.

\begin{takeawaybox}
  \textbf{Takeaway \#2:} Training on context lengths similar to evaluation benchmarks outperforms training on longer contexts by 1.4--3.0 points VA.
  \end{takeawaybox}

\subsubsection{Page indices}
\label{sec:page_indices}
Beyond context length considerations, we identify another practical intervention: explicit page numbering. Referencing pages by number is a desirable property for a long-document VLM, where you may wish to focus the model on a specific page or set of pages. Motivated by this, we measure the impact of prepending a page index to each image in context. We consider two settings: (i) during training and evaluation vs neither and (ii) during evaluation only vs neither. The results, shown in Table \ref{tab:page_indices}, show that page indices are a strong boost to VA performance and improve scores on MMLBD-C by a significant margin. We also find that including page indices during training is necessary to improve VA with page indices during evaluation.

\begin{table}[t]
  \begin{center}
  \resizebox{\textwidth}{!}{%
  \begin{tabular}{cccccccccc}
  \toprule
  \multicolumn{2}{c}{\bf Page Indices} & \multicolumn{1}{c}{\bf VA} & \multicolumn{1}{c}{\bf LCA} & \multicolumn{1}{c}{\bf MMLBD-C} & \multicolumn{1}{c}{\bf MMLB 128K} & \multicolumn{1}{c}{\bf SlideVQA} & \multicolumn{1}{c}{\bf Helmet} & \multicolumn{1}{c}{\bf LongBench v2} & \multicolumn{1}{c}{\bf DUDE} \\
  \cmidrule(lr){1-2}
  \bf Train & \bf Eval & & & & & & & & \\
  \midrule
  \checkmark & \checkmark & 84.0 \textcolor{teal}{(+2.8)} & 85.7 \textcolor{teal}{(+1.1)} & 45.1 \textcolor{teal}{(+2.8)} & 69.2 \textcolor{teal}{(+4.3)} & 69.4 \textcolor{teal}{(+1.0)} & 64.7 \textcolor{red}{(-1.4)} & 43.0 \textcolor{red}{(-3.0)} & 52.7 \textcolor{red}{(-0.5)} \\
  $\times$ & $\times$ & 81.2 & 84.6 & 42.3 & 64.9 & 68.4 & 66.1 & 46.0 & 53.2 \\
  \midrule
  $\times$ & $\times$ & 84.9 & 83.0 & 47.4 & 65.7 & 71.2 & 53.1 & 38.0 & 54.0 \\
  $\times$ & \checkmark & 83.9 \textcolor{red}{(-1.0)} & 82.7 \textcolor{red}{(-0.3)} & 48.2 \textcolor{teal}{(+0.8)} & 62.5 \textcolor{red}{(-3.2)} & 69.7 \textcolor{red}{(-1.5)} & 52.6 \textcolor{red}{(-0.4)} & 40.0 \textcolor{teal}{(+2.0)} & 53.9 \textcolor{red}{(-0.1)} \\
  \bottomrule
  \end{tabular}
  }%
  \end{center}
  \caption{Page indices during training and evaluation vs neither and during evaluation only vs neither. We see that including page indices during training is important to benefit from page indices during evaluation.}\label{tab:page_indices}
\end{table}

\begin{takeawaybox}
\textbf{Takeaway \#3:} Adding explicit page indices during both training and evaluation provides a simple, high-impact boost: +2.8 points on MMLBD-C and +2.8 points on visual LC average.
\end{takeawaybox}

\subsection{Preference optimization (LongPO)}
\label{sec:longpo}
While SFT provides strong improvements, preference optimization offers an alternative paradigm for aligning model behavior. In an effort to build the strongest possible visual long-document model, we train Qwen3 VL using \nameref{pa:longpo}. Rather than applying this in the self-improvement setting, we use the stronger 235B model for answer generation. We use the same training settings as recommended by \citep{longpo} and train on 36K examples.

We find that LongPO is a strong improvement over SFT on VA and improves Qwen3 VL's scores on MMLBD-C, matching the performance of Qwen3 VL 235B A22B. We include these results along with the best SFT results for Qwen3 VL and Mistral and the baseline models in Table \ref{tab:best_ckpts}.


\begin{table}[t]
  \begin{center}
  \resizebox{\textwidth}{!}{%
  \begin{tabular}{lcccccccc}
  \toprule
  \multicolumn{1}{l}{\bf Checkpoint} & \multicolumn{1}{c}{\bf VA} & \multicolumn{1}{c}{\bf LCA} & \multicolumn{1}{c}{\bf MMLBD-C} & \multicolumn{1}{c}{\bf MMLB 128K} & \multicolumn{1}{c}{\bf SlideVQA} & \multicolumn{1}{c}{\bf Helmet} & \multicolumn{1}{c}{\bf LongBench v2} & \multicolumn{1}{c}{\bf DUDE} \\
  \midrule
  \textbf{Qwen3 VL 235B A22B} & \large\textbf{99.0} & \large\textbf{99.2} & 56.2 & \large\textbf{78.6} & \large\textbf{84.5} & \large\textbf{67.6} & \large\textbf{50.0} & 59.1 \\
  LongPO Short Stage & 94.6 \textcolor{teal}{(+0.4)} & 93.1 \textcolor{teal}{(+0.3)} & 56.4 \textcolor{teal}{(+2.6)} & 75.6 \textcolor{teal}{(+5.2)} & 75.5 \textcolor{red}{(-1.7)} & 62.9 \textcolor{red}{(-0.1)} & 42.0 & 56.0 \textcolor{red}{(-5.8)} \\
  \textbf{Qwen3 VL 32B} & 94.2 & 92.8 & 53.8 & 70.4 & 77.2 & 63.0 & 42.0 & \large\textbf{61.8} \\
  Qwen3 VL Plain Distillation Short Stage & 92.5 \textcolor{red}{(-1.6)} & 92.5 \textcolor{red}{(-0.2)} & \large\textbf{57.3} \textcolor{teal}{(+3.5)} & 73.8 \textcolor{teal}{(+3.4)} & 66.8 \textcolor{red}{(-10.4)} & 65.7 \textcolor{teal}{(+2.7)} & 44.0 \textcolor{teal}{(+2.0)} & 54.8 \textcolor{red}{(-7.0)} \\
  \midrule
  Mistral Plain Distillation* & 84.9 \textcolor{teal}{(+4.3)} & 83.0 \textcolor{teal}{(+5.9)} & 47.4 \textcolor{teal}{(+6.1)} & 65.7 \textcolor{red}{(-0.7)} & 71.2 \textcolor{teal}{(+3.4)} & 53.1 \textcolor{teal}{(+16.0)} & 38.0 \textcolor{red}{(-1.0)} & 54.0 \textcolor{teal}{(+1.2)} \\
  \textbf{Mistral 3.1 Small (24B)} & 80.2 & 76.7 & 41.4 & 66.4 & 67.8 & 37.0 & 39.0 & 52.8 \\
  \bottomrule
  \end{tabular}
  }%
  \end{center}
  \caption{Summary of best performing checkpoints: results for LongPO, top SFT checkpoints and baseline models with deltas shown relative to the base model. While not shown in the table, we report SOTA values: Qwen3-VL 32B plain distillation reaches SOTA performance on MMLBD-C and matches Qwen3 VL 235B A22B on MMLongBenchDoc with an accuracy of 56.3 (Qwen3 VL at 52.6) vs 56.7. For models under 32B, the Mistral checkpoint outperforms GLM 4.1V Thinking 9B with an accuracy of 46.8 (Mistral at 40.8) vs 42.4}\label{tab:best_ckpts}
\end{table}

\begin{takeawaybox}
\textbf{Takeaway \#4:} LongPO provides stronger improvements than SFT on VA (+2.1), though it requires more than 2$\times$ the compute (see \hyperref[pa:longpo]{LongPO background}). For best MMLBD-C performance specifically, plain distillation with SFT is more effective.
\end{takeawaybox}

\subsection{Self-improvement}
\label{sec:selfimprove}
The methods above assume access to a stronger teacher model for distillation. However, in the case of frontier models, no stronger teacher is available. Thus, to advance the SOTA, our options include expensive human annotation and methods for increasing LC performance in a self-improving manner. While existing work on preference optimization for short-to-long extensions have shown strong results \citep{longpo,solopo} in this setting, the chosen responses for these methods are generated from a localized subset of the full context, where the question originates, while the rest of the context is treated as irrelevant regardless of the content.

Our proposed \hyperref[sec:answer_generation]{recursive answer generation pipeline} is not limited to the context available at the time of question generation and thus we distill an algorithm into the model that involves a non-trivial search over the full context. Our recursive pipeline is compatible with LongPO and SoLoPO, but here we focus on self-improvement with SFT. We also note that our CPT tasks are within the self-improving setting as we used Mistral in the construction of the CPT data. We showcase two SFT checkpoints and the CPT scores in Table~\ref{tab:self_improvement} and find that CPT for 100B tokens achieves the strongest self-improvement performance with +3.8 VA and though SFT uses far less compute, we find that SFT alone is also effective, yielding +3.2 VA, while also surpassing CPT on MMLBD-C.

\begin{table}[t]
  \begin{center}
  \resizebox{\textwidth}{!}{%
  \begin{tabular}{lcccccccc}
  \toprule
  \multicolumn{1}{l}{\bf Checkpoint} & \multicolumn{1}{c}{\bf VA} & \multicolumn{1}{c}{\bf LCA} & \multicolumn{1}{c}{\bf MMLBD-C} & \multicolumn{1}{c}{\bf MMLB 128K} & \multicolumn{1}{c}{\bf SlideVQA} & \multicolumn{1}{c}{\bf Helmet} & \multicolumn{1}{c}{\bf LongBench v2} & \multicolumn{1}{c}{\bf DUDE} \\
  \midrule
  Self-Improvement - SFT Only & 83.8 \textcolor{teal}{(+3.2)} & 77.6 \textcolor{teal}{(+0.5)} & 45.2 \textcolor{teal}{(+3.8)} & 68.7 \textcolor{teal}{(+2.3)} & 69.2 \textcolor{teal}{(+1.4)} & 32.4 \textcolor{red}{(-4.7)} & 35.0 \textcolor{red}{(-4.0)} & 52.7 \textcolor{red}{(-0.1)} \\
  Self-Improvement - Instruct + CPT & 82.9 \textcolor{teal}{(+2.3)} & 79.8 \textcolor{teal}{(+2.7)} & 43.3 \textcolor{teal}{(+2.0)} & 70.0 \textcolor{teal}{(+3.6)} & 64.9 \textcolor{red}{(-2.9)} & 42.3 \textcolor{teal}{(+5.3)} & 39.0 & 56.8 \textcolor{teal}{(+4.0)} \\
  \midrule
  CPT (100B) & 84.4 \textcolor{teal}{(+3.8)} & 84.6 \textcolor{teal}{(+7.5)} & 42.7 \textcolor{teal}{(+1.3)} & 69.3 \textcolor{teal}{(+2.9)} & 68.0 \textcolor{teal}{(+0.2)} & 51.7 \textcolor{teal}{(+14.6)} & 47.0 \textcolor{teal}{(+8.0)} & 60.1 \textcolor{teal}{(+7.3)} \\
  \bottomrule
  \end{tabular}
  }%
  \end{center}
  \caption{Results for self-improvement with recursive answer generation pipeline. Instruct + CPT trains from the Instruct model merged with the CPT vector. For details on CPT, see Appendix \ref{sec:cpt_ablations}.}\label{tab:self_improvement}
\end{table}

\begin{takeawaybox}
\textbf{Takeaway \#5:} Our recursive answer generation pipeline enables self-improvement, achieving +3.2--3.8 points on visual LC average through CPT or SFT alone.
\end{takeawaybox}

\section{Conclusion}
\label{sec:conclusion}


Our study, spanning CPT, SFT and preference optimization at scales up to 100B tokens on 24B and 32B parameter models yields several actionable insights. First, \emph{CPT is not always necessary}: when the base model's context length is sufficient, SFT or LongPO alone achieve competitive visual long-document performance, though CPT improves LCA. Second, \emph{matching training context to evaluation context outperforms training on longer contexts}. Third, simple interventions like \emph{page indices} provide substantial gains (+2.8 points on MMLBD-C) with minimal implementation effort. Fourth, we demonstrate that \emph{visual LC training transfers to LC text performance} (+11.5 points on HELMET). Finally, our synthetic data pipelines enable \emph{self-improvement}, demonstrating their capability for weak-to-strong LC performance.

\paragraph{Limitations and future work.}
Our evaluation suite, while comprehensive, under-represents extreme-length documents (most benchmarks are under 128K tokens), limiting our ability to verify performance at the full 344K context length. The interaction between CPT and SFT remains incompletely understood: they do not compose additively across many benchmarks, suggesting opportunities for mixed-stage training or replay mechanisms.

\paragraph{Conclusion.}
We provide open, reproducible recipes for training long-context visual document models that achieve state-of-the-art performance. Beyond the recipes themselves, we quantify compute/data trade-offs and identify high-impact techniques (page indices, context length matching) that practitioners can adopt immediately. We release \textsc{MMLBD-C} to improve evaluation quality and hope our findings accelerate progress in long-document understanding.

\section*{Acknowledgements}
We thank Oskar Hallstr\"om for his suggestions on model merging and valuable assistance with experiment design. We also thank the LightOn team for their support and feedback throughout the project.

This work was granted access to the HPC resources of IDRIS under the allocations \texttt{AS011016449} and \texttt{A0181016214} made by GENCI enabling us to use the Jean Zay supercomputer. We thank the IDRIS support team for their valuable help.

This project also received funding from the BPI Scribe project.

We acknowledge EuroHPC JU for awarding the project ID EHPC-AIF-2025FL01-523 access to MareNostrum5 at BSC, Spain.

\bibliography{paper_refs}
\bibliographystyle{colm2026_conference}

\clearpage
\appendix
\section{Appendix}
\label{sec:appendix}

\etocsettocstyle{\subsection*{Appendix Contents}}{}
\localtableofcontents

\clearpage
\subsection{Reproducibility statement}
\label{sec:repro}
\paragraph{Recipes.}
We publish an html file for easy exploration and reproducibility of our training runs. This includes all benchmark scores, training method, base model, merging strategy and data composition. This along with the synthetic data pipelines are available on GitHub \footnote{\url{https://huggingface.co/collections/lightonai/orion}, \url{https://github.com/lightonai/distilabel/tree/lc_sft_pipelines}}.

\paragraph{Models.}
We apply CPT to Mistral Base and Qwen3 VL Instruct since the base model is not available. SFT and LongPO are applied from the instruct checkpoint or the instruct checkpoint merged with the CPT vector. For Mistral, we extend the context length to $344\text{K}$ tokens and for Qwen3~VL, we simply maintain the original context length of $256\text{K}$ tokens.

\paragraph{Training details.}
\label{pa:training_details}
For all training, we use the AdamW \citet{adamw} optimizer with $\epsilon = 10^{-9}$. We use sequence packing, avoid truncating sequences, and normalize loss by the total number of assistant tokens. We dynamically scale document resolution when it will not fit entirely within the context, varying the maximum side resolution from $616$ to $840$ for CPT and from $728$ to $1400$ for SFT/LongPO. For Mistral, the effective patch size is $28$, so an $840 \times 840$ image corresponds to $(840 / 28)^2 = 900$ tokens. For Qwen3~VL, the effective patch size is $32$.

For Mistral, stage~1 forms packed sequences of $128\text{K}$ tokens and stage~2 is $336\text{K}$ tokens. For Qwen3~VL, stage~1 is $128\text{K}$ tokens and stage~2 is $256\text{K}$ tokens. We use ring attention for sequence parallelism (SP) \citet{ring_attn,ring_flash_attn}. Optimizer hyperparameters are summarized in Table~\ref{tab:training_hyperparams} and parallelism configurations in Table~\ref{tab:parallelism_config}. Mistral's RoPE \citet{rope} $\theta$ is already set to $10^9$ so we do not increase it. For Qwen3~VL, we do not extend the context length so we do not modify the RoPE $\theta$.

\begin{table}[h]
  \begin{center}
  \resizebox{\textwidth}{!}{%
  \begin{tabular}{lccccccc}
  \toprule
  \multicolumn{1}{l}{\bf Phase} & \multicolumn{1}{c}{\bf Schedule} & \multicolumn{1}{c}{\bf Max LR} & \multicolumn{1}{c}{\bf Warmup/Decay} & \multicolumn{1}{c}{\bf $\beta_1$} & \multicolumn{1}{c}{\bf $\beta_2$} & \multicolumn{1}{c}{\bf Weight Decay} & \multicolumn{1}{c}{\bf Grad Clip} \\
  \midrule
  CPT & Cosine & $4e\text{-}6$ & 10\% tokens & 0.9 & 0.999 & 0.1 & 1.0 \\
  SFT & WSD \citet{wsd} & $5e\text{-}6$ & 10\% samples & 0.9 & 0.999 & 0.1 & 1.0 \\
  LongPO & WSD \citet{wsd} & $5e\text{-}7$ & 10\% samples & 0.9 & 0.99 & 0.0 & 1.0 \\
  \bottomrule
  \end{tabular}
  }%
  \end{center}
  \caption{Optimizer hyperparameters for each training phase. For LongPO, we additionally use $\beta=0.1$ and $\lambda=0.01$ from Eq.~\ref{eq:longpo_objective}.}\label{tab:training_hyperparams}
\end{table}

\begin{table}[h]
  \begin{center}
  \resizebox{\textwidth}{!}{%
  \begin{tabular}{llcccccc}
  \toprule
  \multicolumn{1}{l}{\bf Phase} & \multicolumn{1}{l}{\bf Stage} & \multicolumn{1}{c}{\bf Hardware} & \multicolumn{1}{c}{\bf SP} & \multicolumn{1}{c}{\bf DP [shard, replicate]} & \multicolumn{1}{c}{\bf Batch Size} & \multicolumn{1}{c}{\bf Tokens/Batch} \\
  \midrule
  CPT & Stage 1 & H100 & 16 & [16, 6] & 6 & 768K \\
  CPT & Stage 2 & H100 & 48 & [16, 6] & 2 & 672K \\
  \midrule
  SFT & Stage 1 & H100 & 16 & [16, 3] & 3 & 768K \\
  SFT & Stage 2 & H100 & 48 & [16, 6] & 2 & 672K/512K \\
  \midrule
  LongPO & Stage 1 & H100 & 48 & [48, 1] & 1 & 128K \\
  LongPO & Stage 2 & H200 & 24 & [24, 1] & 1 & 512K \\
  \bottomrule
  \end{tabular}
  }%
  \end{center}
  \caption{Parallelism and hardware configuration for each training phase and stage. SP = sequence parallelism degree, DP = data parallelism.}\label{tab:parallelism_config}
\end{table}

\paragraph{Model merging.}
For Mistral CPT, we train from the base model and merge the CPT vector into the instruct model with a scaling factor of 0.5. Generally, for all other training types, SFT and LongPO, and for Qwen3 VL we use a scaling factor of 0.25 on the training vector. Specifics for each checkpoint are detailed in the leaderboard.

\paragraph{Recipes used in Figure~\ref{fig:param_vs_mmlbdc}.}
\label{pa:main_recipes}
In Figure~\ref{fig:param_vs_mmlbdc}, we show the highest performing checkpoints on MMLBD-C. More precisely, here are the recipes used for each checkpoint:
\begin{itemize}
  \item Mistral CPT 100B: Length curriculum.
  \item Self-improving: training from Mistral Instruct with the recursive and distractors short pipelines (see Appendix~\ref{sec:appendix_alternative_synthetic_data_pipelines}) without \hyperref[sec:external_sft_composition]{external SFT} data. 2.7K examples from distractors short, 21.5K examples from recursive, 250 examples from unanswerable, 111 examples from multi-turn.
  \item Mistral SFT 'Distill': 50K examples with Magpie questions with plain distillation from Qwen3 VL 235B A22B. Also including external SFT data. 50K examples from Magpie + plain distillation, 500 examples from multi-turn, 10K examples from Luth \citet{luthsft}, 10K examples from Smoltalk2 \citet{smoltalk2}, 1K examples of multi-page OCR on PDFA, 1K examples from DocFinQA \citet{docfinqa}, 2K examples from ChartQA \citet{chartqa} adapted to multi-page.
  \item Qwen3 VL CPT 10B: length difficulty curriculum.
  \item Qwen3 VL SFT 'Distill': 50K examples, SP and MP questions with the same external SFT data as Mistral SFT 'Distill'.
  \item LongPO 'Distill': 35K examples, SP and MP questions with plain distill answers from Qwen3 VL 235B A22B.
\end{itemize}

You can find the distribution of examples used to create our Luth and Smoltalk2 datasets in Appendix~\ref{sec:external_sft_composition}.

\subsection{Corpus details}
\label{sec:corpus_details}
To construct the search queries, we begin with broad categories: arxiv topics, energy industry, financial, government and artificial intelligence. These topics are recursively expanded to form a large set of specific queries. PDFs are then retrieved, deduplicated and filtered for renderability and maximum length. We additionally translate all queries to French and gather an equal French set. During synthetic data generation, we use the question generator to filter these for pages that have more than 100 words, are not table of contents or bibliographies and have content suitable for questions.

\paragraph{Hard negatives.}
We use an in-house DSE \citet{dse} model to mine hard negatives from page embeddings. For each page we store the top 128 most similar pages. We use these to construct challenging examples with distracting pages or similar pages across multiple documents, simulating RAG scenarios.

\begin{figure}[t]
\begin{center}
\includegraphics[width=0.48\linewidth]{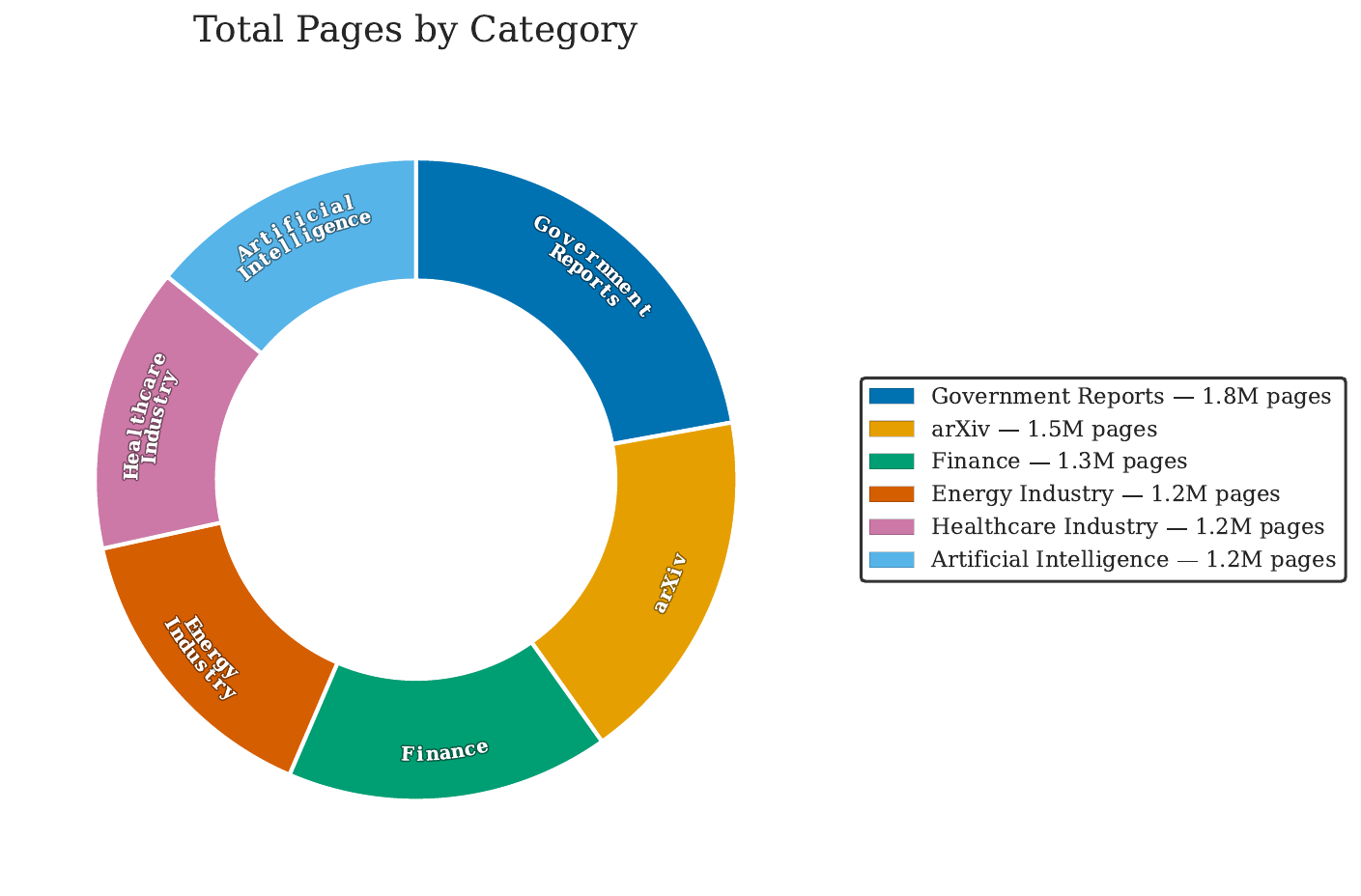}\hfill
\includegraphics[width=0.48\linewidth]{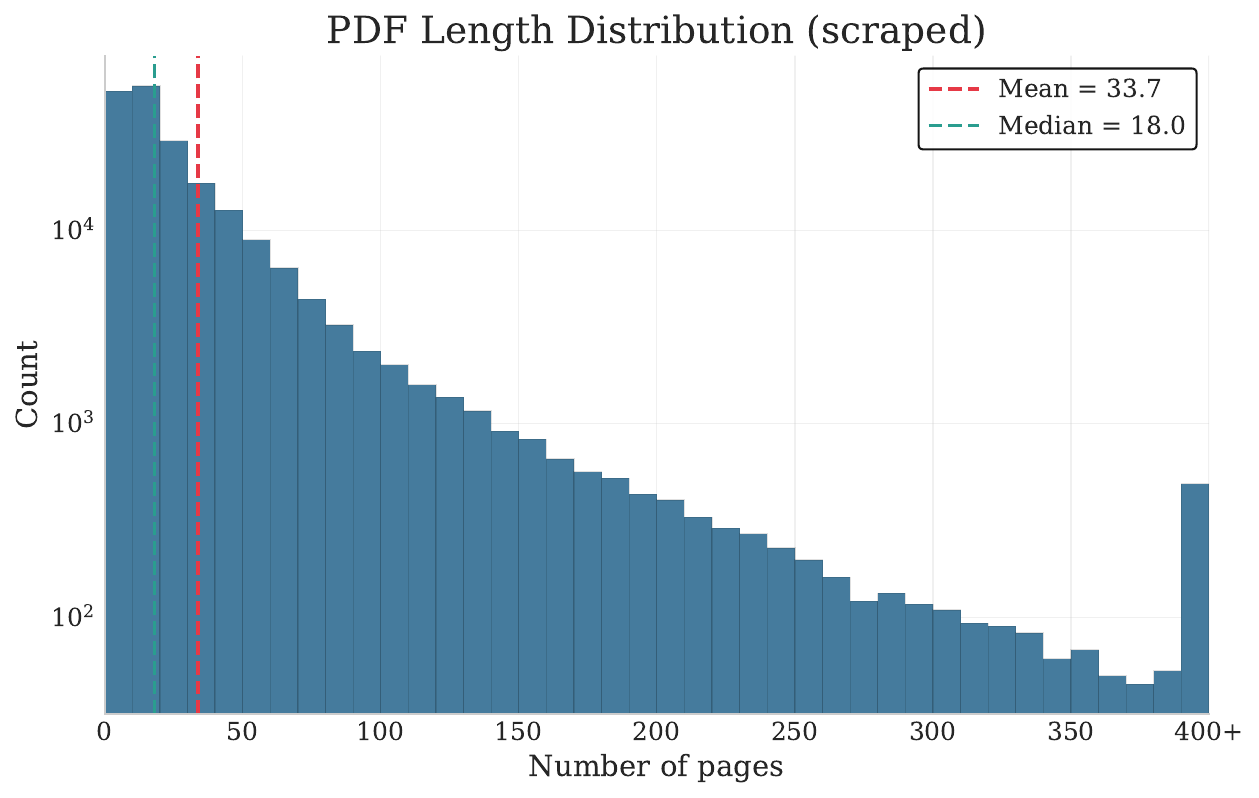}
\end{center}
\caption{Overview of the scraped PDF corpus: (left) total pages by top-level category (categories are recursively refined to generate search queries); (right) distribution of number of pages per PDF.}
\label{fig:scraped_corpus_overview}
\end{figure}

\begin{figure}[t]
\begin{center}
\includegraphics[width=0.48\linewidth]{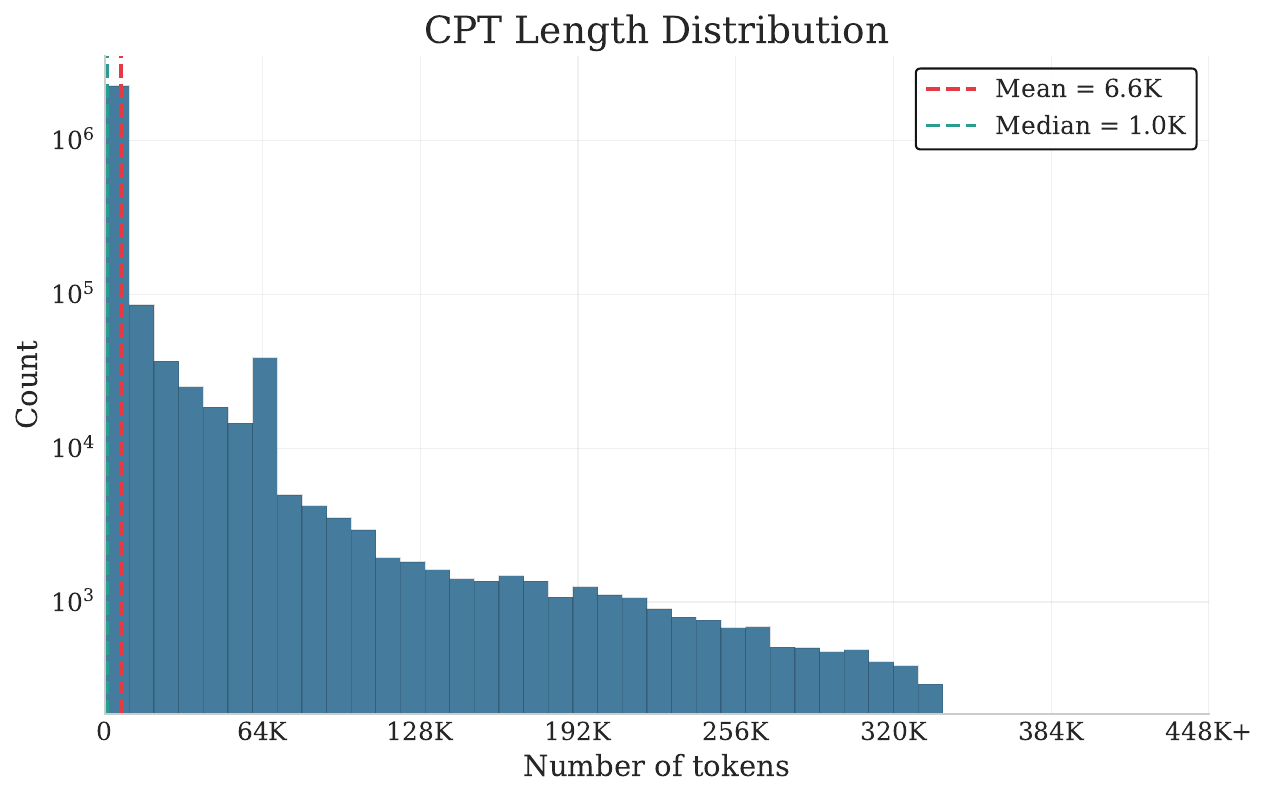}\hfill
\includegraphics[width=0.48\linewidth]{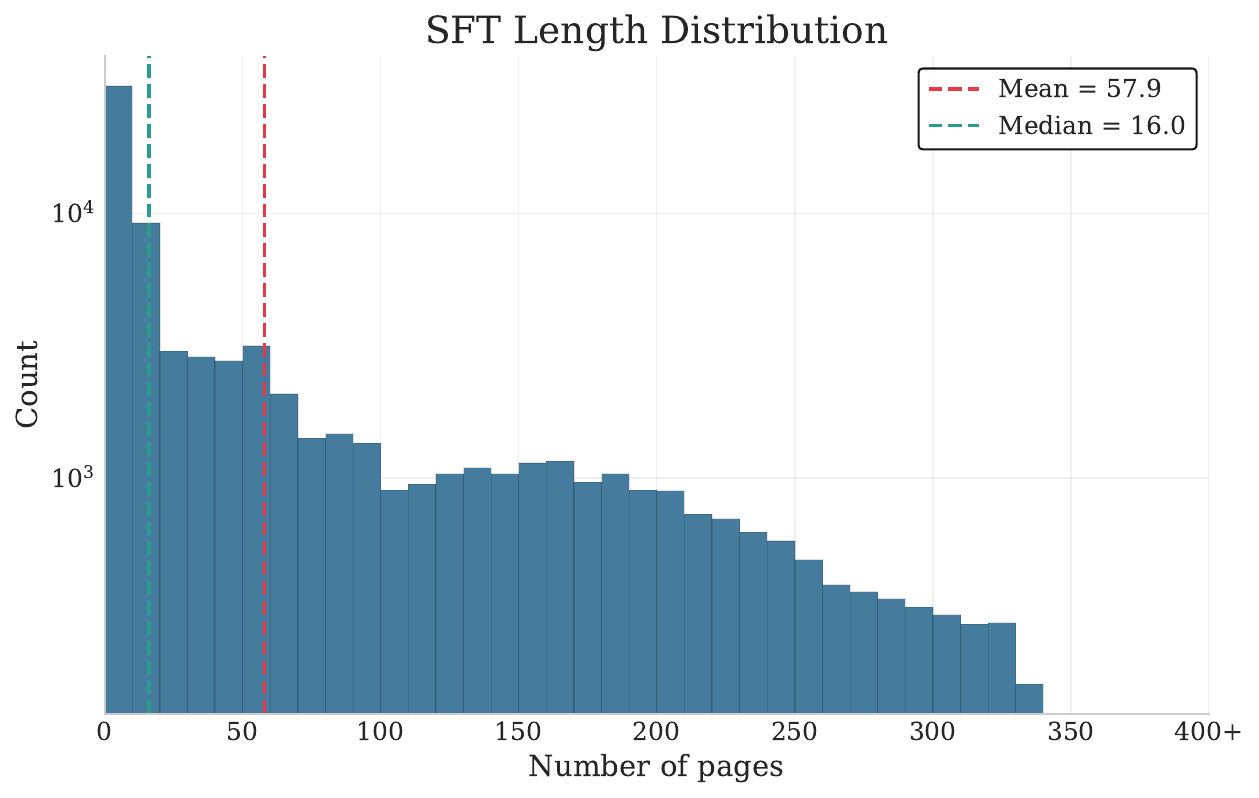}
\end{center}
\caption{Length distributions of training examples. (Left) CPT example length (tokens): image tokens are estimated as 1024 tokens per page; text-only samples shorter than 1024 tokens are clipped to 1024. Note that the LC text data from Prolong is very strongly skewed towards short examples. (Right) SFT example length (pages).}
\label{fig:train_length_dists}
\end{figure}

\begin{figure}[t]
\begin{center}
\includegraphics[width=0.6\linewidth]{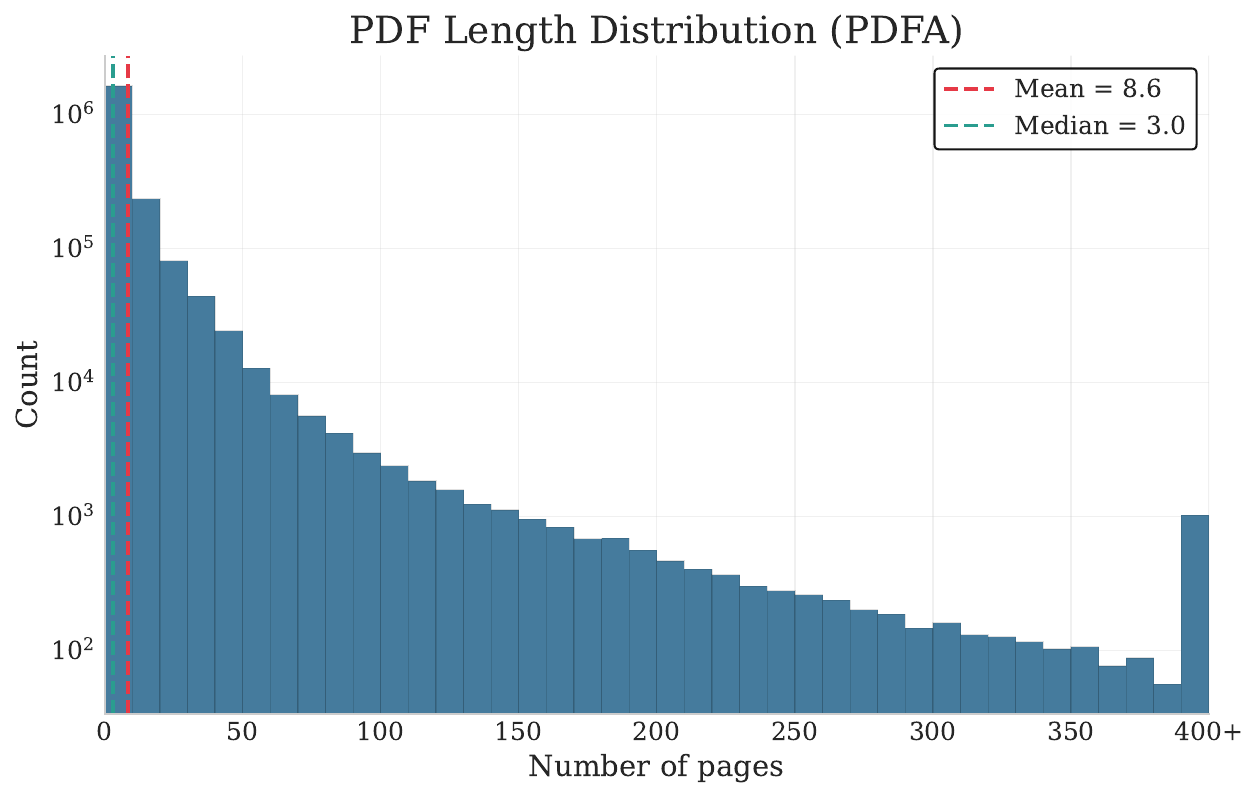}
\end{center}
\caption{Distribution of number of pages per PDF in the PDFA English split.}
\label{fig:pdf_length_pdfa}
\end{figure}

\begin{figure}[t]
\begin{center}
\includegraphics[width=0.75\linewidth]{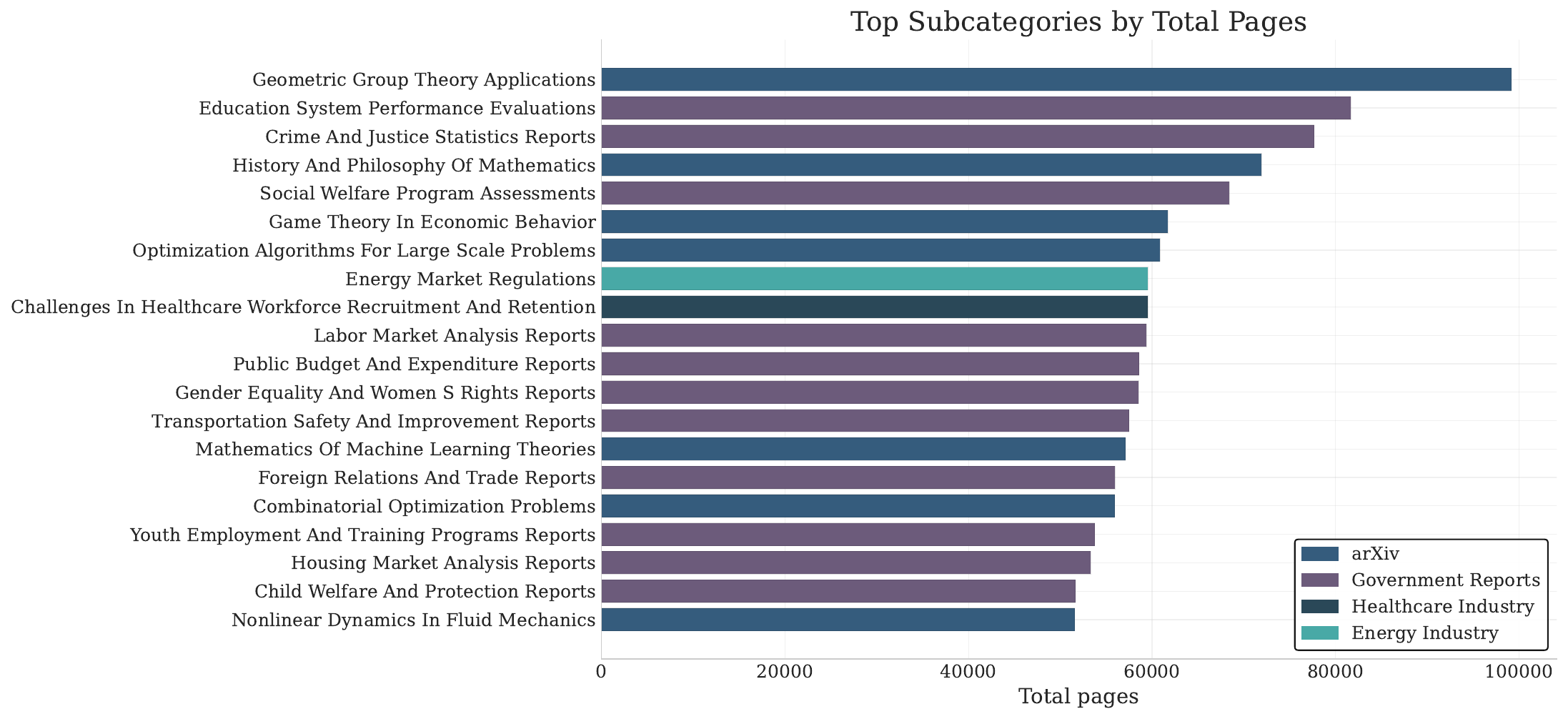}
\end{center}
\caption{Top subcategories by total pages within the scraped PDF corpus (grouped by parent category).}
\label{fig:top_subcats_pages}
\end{figure}

\subsection{Evaluation}
\label{sec:evaluation_benchmarks}

We evaluate on a suite of long-context benchmarks spanning document VQA and long-context text tasks, along with a few knowledge and reasoning benchmarks to measure degradation. Specifically, we include MMLongBenchDoc \citet{mmlbd} (and our corrected variant \textsc{MMLBD-C}); MMLongBench \citet{mmlb} at 32K and 128K context (document QA, visual RAG, ICL, summarization); SpiQA \citet{spiqa}; SlideVQA Mini \citet{slidevqa}; HELMET \citet{helmet} at 32K and 128K context (recall, RAG, summarization, ICL, reranking); LongBench v2 \citet{longbench_v2}; DUDE Mini \citet{dude}; TableVQA \citet{tablevqa}; MMMU-Pro \citet{mmmupro}; TinyMMLU \citet{tinybenchmarks,mmlu}; MM-MT \citet{mmmt}; GPQA \citet{gpqa}; TinyGSM8K \citet{tinybenchmarks,gsm8k}; Internal single-page QA; Internal multi-page QA; Internal multi-page QA with hard negatives. In contrast to the default VLM Eval Kit \citet{vlmevalkit} settings, we increase the maximum number of pages from 120 to 336 for MMLongBenchDoc and \textsc{MMLBD-C}, and we set the maximum resolution to $1024 \times 1024$ to ensure long examples fit in context while preserving fine details. We list the specific metrics used for each benchmark below. Due to the large number of evaluations, we limit expensive benchmarks (HELMET and MMLongBench) to 20 samples per task, and we use a local judge, selecting GLM 4.5V due to its strong performance on MMLongBenchDoc and LM Arena \citet{lmarena} while being outside the model families we train to avoid self-preference bias \citet{selfpreference}.

Table~\ref{tab:benchmark_metrics} lists the primary metric used for each benchmark in our evaluation suite. We release an html file with the full set of scores for easy exploration.

\begin{table}[t]
  \centering
  \resizebox{\textwidth}{!}{%
  \begin{tabular}{ll}
  \toprule
  \textbf{Benchmark} & \textbf{Metric} \\
  \midrule
  MMLongBenchDoc / MMLBD-C & F1 (overall\_f1) \\
  MMLongBench (32K/128K) & Avg of task-specific metrics$^*$ \\
  SpiQA & LLM Judge Score (0--5, normalized to 0--100) \\
  SlideVQA Mini & ANLS (Average Normalized Levenshtein Similarity) \\
  HELMET (32K/128K) & Overall Score \\
  LongBench v2 & Overall Accuracy \\
  DUDE Mini & ANLS (Average Normalized Levenshtein Similarity) \\
  TableVQA & Average Accuracy across subtasks \\
  MMMU-Pro & Overall Accuracy \\
  TinyMMLU & Normalized Accuracy (acc\_norm) \\
  MM-MT & LLM Judge Score (0--10, normalized to 0--100) \\
  GPQA & Exact Match (flexible extraction) \\
  TinyGSM8K & Exact Match (flexible extraction) \\
  Internal benchmarks & Average LLM Judge Score (1--5, normalized to 0--100) \\
  \bottomrule
  \end{tabular}
  }%
  \caption{Primary metrics used for each benchmark. All scores are normalized to 0--100 before averaging. $^*$MMLongBench task-specific metrics: Visual RAG (infoseek, viquae): sub\_em, ICL (cars196, food101, inat2021, sun397): cls\_acc, Summarization (gov-report, multi-lexsum): judge\_f1.}
  \label{tab:benchmark_metrics}
\end{table}

\begin{figure}[t]
  \centering
  \begin{subfigure}[t]{0.48\linewidth}
      \centering
      \includegraphics[width=\linewidth]{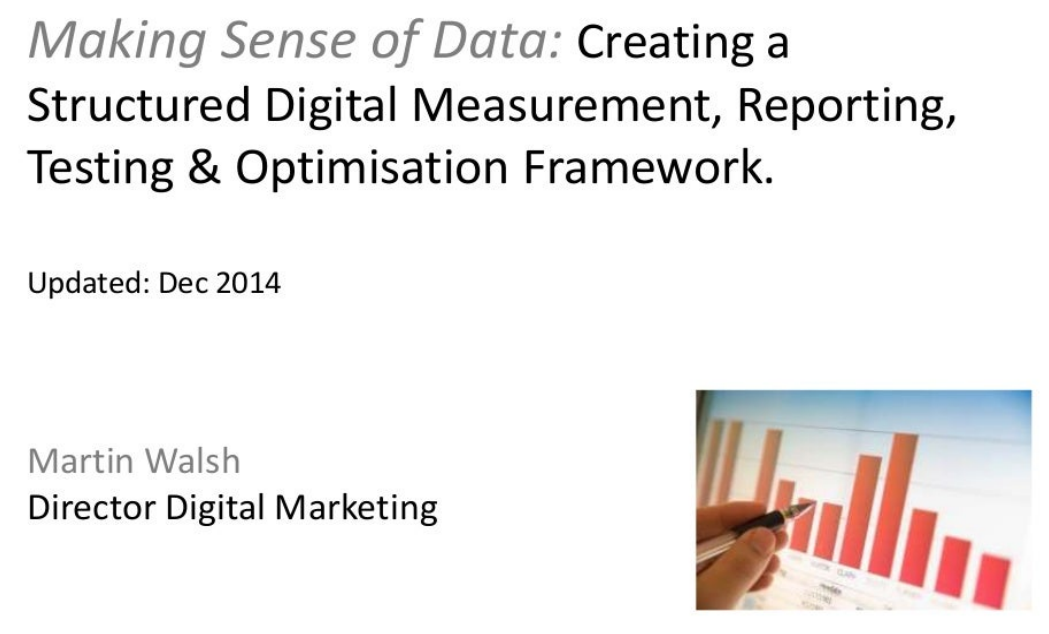}
      \caption{Document mismatch}
      \label{fig:mmlbdc_mismatch}
  \end{subfigure}
  \hfill
  \begin{subfigure}[t]{0.48\linewidth}
      \centering
      \includegraphics[width=\linewidth]{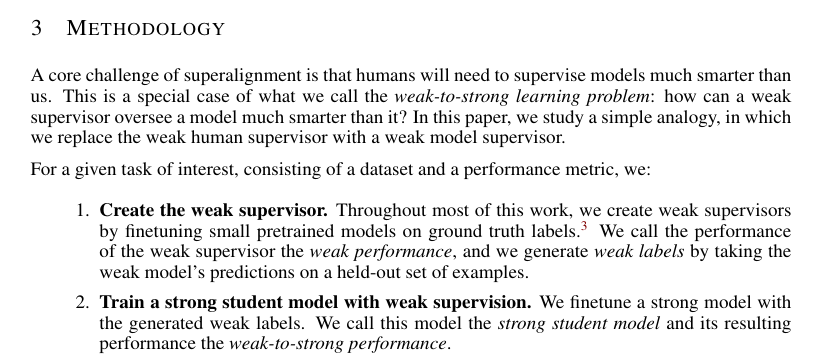}
      \caption{Underspecified question}
      \label{fig:mmlbdc_underspecified}
  \end{subfigure}

  \vspace{0.5em}
  \begin{subfigure}[t]{0.48\linewidth}
      \centering
      \includegraphics[width=\linewidth]{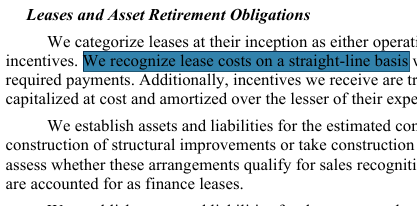}
      \caption{Typo: ``least'' $\to$ ``lease''}
      \label{fig:mmlbdc_typo}
  \end{subfigure}
  \hfill
  \begin{subfigure}[t]{0.48\linewidth}
      \centering
      \includegraphics[width=\linewidth]{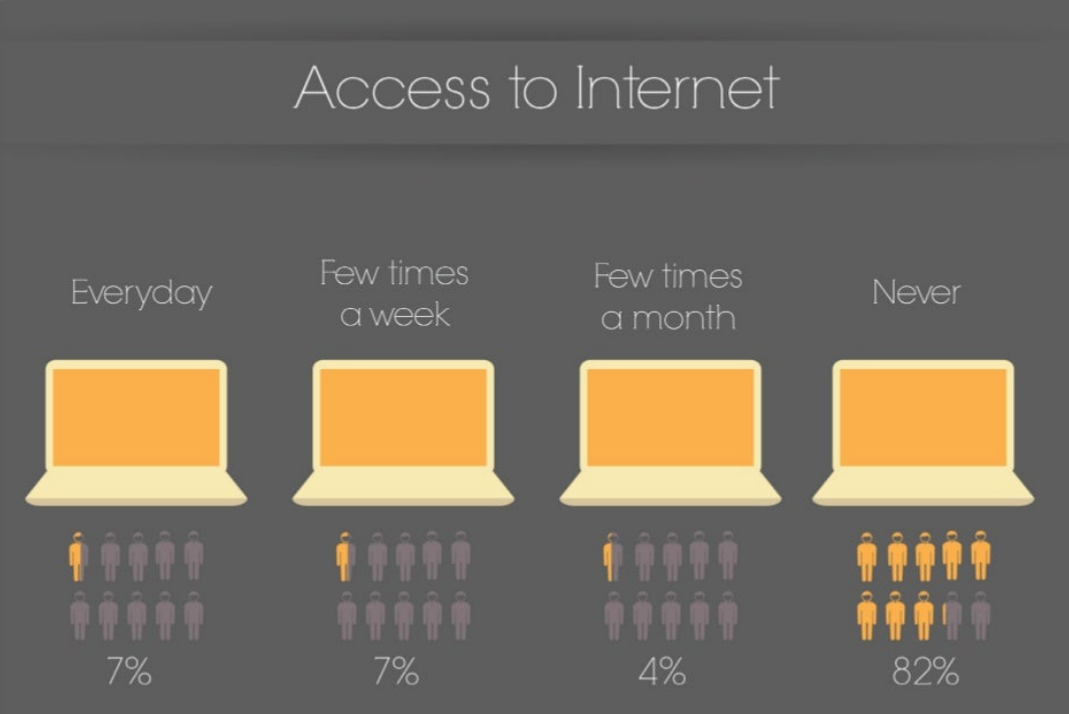}
      \caption{Incorrect ``Not answerable''}
      \label{fig:mmlbdc_incorrect}
  \end{subfigure}
  \caption{Examples of issues in MMLongBenchDoc: (a)~question paired with wrong document (``List all the PM health effects that increse by more than 35\% in India and Thailand.''), (b)~underspecified question (``List all the sections that discuss about the experiment setup?'' $\to$ ``['Section 4.1', 'Section 4.2', 'Section 4.3', 'Appendix A']''), the answer does not include the Methodology section which discusses the experiment, (c)~typo causing confusion (``How do Amazon recognize \underline{least} cost?'' $\to$ ``lease cost''), (d)~answerable marked as unanswerable (``How many percentage respondents in this survey access to internet more than two times per month?'' $\to$ ``Not answerable'', however, 7\%+7\%+4\%=18\% access the internet more than two times per month).}
  \label{fig:mmlbdc_examples}
\end{figure}

\begin{figure}[t]
  \begin{center}
  \includegraphics[width=\linewidth]{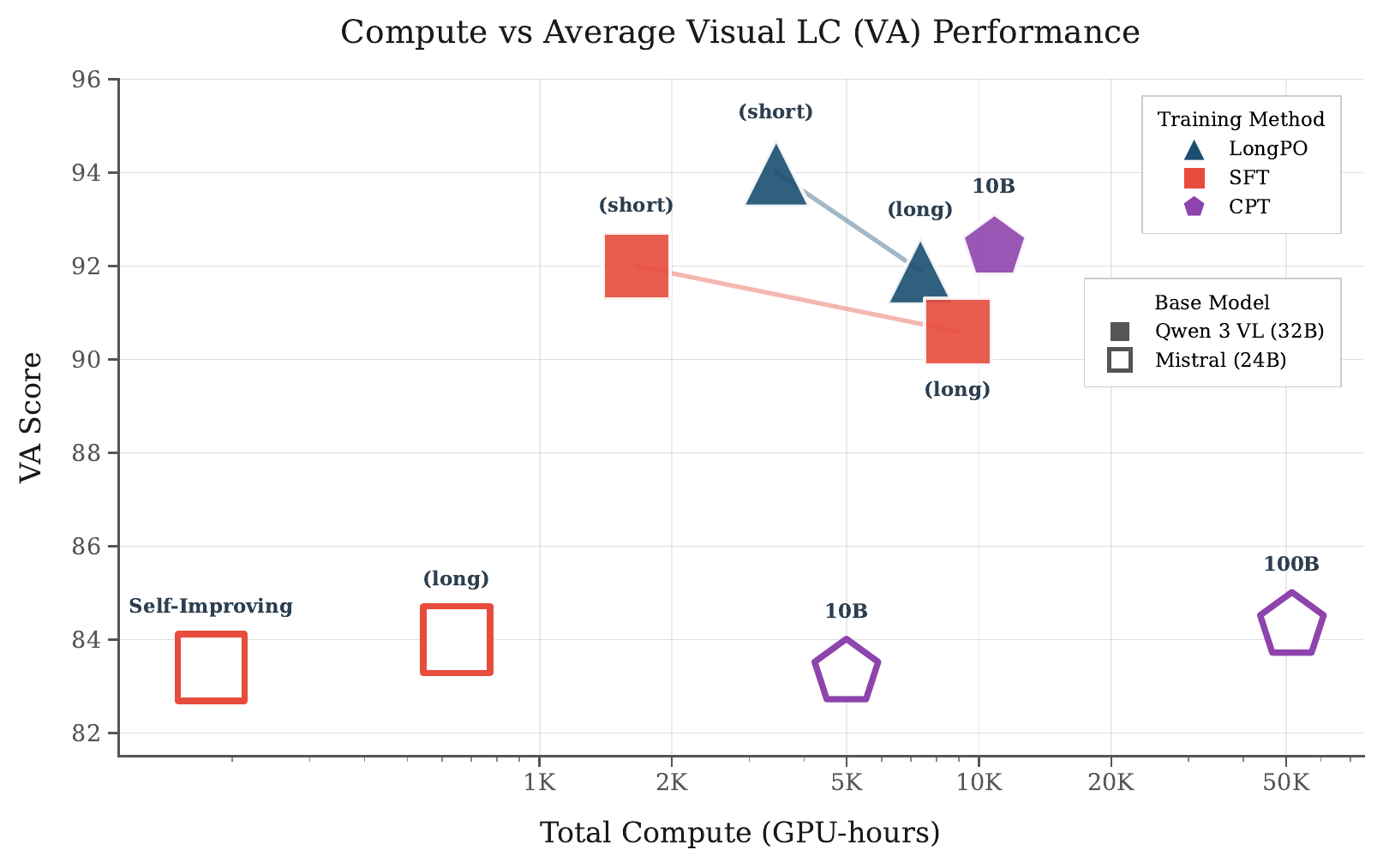}
  \end{center}
  \caption{Compute vs. Average Visual LC (VA) performance across key training runs in this work. For Mistral, we show results for the 'plain distillation' pipeline. For Qwen3 VL, we show SFT and LongPO with answers generated in the same fashion. SFT checkpoints undergo model merging with the CPT vector, so we note that their real total compute is in addition to CPT. Also note that the long stage for LongPO was shortened compared to SFT and we used H200 GPUs. For fair comparison, we scale GPU hours by 2 due to higher memory bandwidth and lower necessary sequence parallelism. This figure shows the degradation when training on context lengths significantly longer than the benchmarks. We also see tradeoffs between CPT, SFT and LongPO in terms of compute.}
  \label{fig:compute_vs_va}
\end{figure}

In Table \ref{tab:eval_variance}, we show results for the variance of VA and LCA across 3 runs. Our aggregate metrics are stable across runs, with $\sigma = 0.33$ for VA and $\sigma = 0.24$ for LCA. However, we note that the variance of MMLongBench is especially high, with $\sigma = 1.66$. This is likely due to limiting the number of examples to 20 per task, with a total of $180$ examples for each context length ($32K$ and $128K$).

\begin{table}[t]
  \begin{center}
  \resizebox{\textwidth}{!}{%
  \begin{tabular}{lcccccccc}
  \toprule
  \multicolumn{1}{l}{\bf Checkpoint} & \multicolumn{1}{c}{\bf VA} & \multicolumn{1}{c}{\bf LCA} & \multicolumn{1}{c}{\bf MMLBD-C} & \multicolumn{1}{c}{\bf MMLB 128K} & \multicolumn{1}{c}{\bf SlideVQA} & \multicolumn{1}{c}{\bf Helmet} & \multicolumn{1}{c}{\bf LongBench v2} & \multicolumn{1}{c}{\bf DUDE} \\
  \midrule
  LongPO Long Stage (Distill) Eval\#3 & 92.8 \textcolor{teal}{(+0.4)} & 91.8 \textcolor{teal}{(+0.3)} & 54.1 \textcolor{teal}{(+0.1)} & 74.4 \textcolor{teal}{(+2.6)} & 74.3 \textcolor{red}{(-0.5)} & 62.2 \textcolor{red}{(-1.2)} & 43.0 \textcolor{teal}{(+1.0)} & 55.5 \textcolor{teal}{(+1.3)} \\
  LongPO Long Stage (Distill) & 92.4 & 91.5 & 54.0 & 71.8 & 74.8 & 63.4 & 42.0 & 54.1 \\
  LongPO Long Stage (Distill) Eval\#2 & 92.0 \textcolor{red}{(-0.4)} & 91.2 \textcolor{red}{(-0.3)} & 53.2 \textcolor{red}{(-0.8)} & 75.8 \textcolor{teal}{(+4.0)} & 73.2 \textcolor{red}{(-1.6)} & 63.2 \textcolor{red}{(-0.2)} & 42.0 & 55.2 \textcolor{teal}{(+1.1)} \\
  \bottomrule
  \end{tabular}
  }%
  \end{center}
  \caption{Evaluation variance across 3 runs.}\label{tab:eval_variance}
\end{table}

\FloatBarrier
\subsection{Page indices format}
\label{sec:page_indices_format}
We prepend a simple page index to each image in the input context. The format is shown below:

\begin{tcolorbox}[colback=gray!5, colframe=gray!50, title=Page Indices Format, fonttitle=\bfseries]
\ttfamily
Page 1:\\
<image>\\
Page 2:\\
<image>\\
Page 3:\\
<image>\\
...
\end{tcolorbox}

This minimal intervention provides explicit positional information that helps the model reference and reason about specific pages in long documents.

\subsection{CPT}
\label{sec:cpt_ablations}
\paragraph{Task Ablations.}
We ablate the impact of each CPT task (see Section~\ref{sec:cpt_tasks}) by comparing the performance of the full CPT mixture vs the performance of the full CPT mixture with one task removed. The scores are shown in Table \ref{tab:cpt_ablations}. We see that removing the FIM task leads to the largest degradation (-3.0 VA) indicating its importance. The impact of the other tasks follows. It is interesting to see Unshuffle with such a large impact; this data requires no model to construct, making it scalable, and targets the model's understanding of the entire document, which is unique among the CPT tasks.

\begin{table}[t]
  \centering
  \begin{tabular}{lc}
  \toprule
  \textbf{Ablation} & \textbf{VA} \\
  \midrule
  Drop Counting & 82.7 \textcolor{red}{(-0.7)} \\
  Drop LC Text & 82.1 \textcolor{red}{(-1.3)} \\
  Drop Key/Position Retrieval & 81.6 \textcolor{red}{(-1.7)} \\
  Drop Unshuffle & 81.2 \textcolor{red}{(-2.1)} \\
  Drop FIM & 80.4 \textcolor{red}{(-3.0)} \\
  \bottomrule
  \end{tabular}
  \caption{CPT task ablations. Each row shows the VA score when one task is removed from the 10B token training set. Deltas are relative to the full mixture (VA = 83.4).}
  \label{tab:cpt_ablations}
\end{table}

\paragraph{Token Distribution.}
Table~\ref{tab:cpt_token_distribution} shows the distribution of tokens across CPT tasks, broken down by training stage.

\begin{table}[t]
  \centering
  \begin{tabular}{lrrr}
  \toprule
  \textbf{Task} & \textbf{Short (B)} & \textbf{Long (B)} & \textbf{Total (B)} \\
  \midrule
  Prolong LC Text & 35.9 & 3.8 & 39.7 \\
  Fill-in-the-Middle & 24.3 & 10.1 & 34.5 \\
  Key/Position Retrieval & 15.2 & 6.3 & 21.5 \\
  Unshuffle & 12.2 & 5.1 & 17.2 \\
  Counting & 2.4 & 0.0 & 2.4 \\
  \midrule
  \textbf{Total} & 90.0 & 25.3 & 115.3 \\
  \bottomrule
  \end{tabular}
  \caption{CPT token distribution by task. Short and Long refer to the first and second training stages respectively.}
  \label{tab:cpt_token_distribution}
\end{table}

\paragraph{Curriculum.}
We study three curriculums:
\begin{itemize}
  \item No curriculum: random order of examples.
  \item Length curriculum: the maximum number of pages seen increases throughout training.
  \item Length-difficulty curriculum: we organize the tasks into the following heuristic order from least difficult to most difficult: LC text $\to$ FIM $\to$ unshuffle $\to$ key/position-based retrieval $\to$ counting, then apply the length curriculum within each task, followed by mixing a portion of examples between tasks.
\end{itemize}

To compare these, we CPT Mistral for 10B tokens on each curriculum and find that the results are similar, with the length curriculum lagging behind in MMLongBenchDoc and MMLBD-C. We scale the training of the length curriculum and the length-difficulty curriculum to 100B tokens and find that both curriculums achieve similar results. We use the length-difficulty curriculum throughout our experiments due to a similar VA score and a higher LCA score.

\begin{table}[t]
  \begin{center}
  \resizebox{\textwidth}{!}{%
  \begin{tabular}{lcccccccc}
  \toprule
  \multicolumn{1}{l}{\bf Checkpoint} & \multicolumn{1}{c}{\bf VA} & \multicolumn{1}{c}{\bf LCA} & \multicolumn{1}{c}{\bf MMLBD-C} & \multicolumn{1}{c}{\bf MMLB 128K} & \multicolumn{1}{c}{\bf SlideVQA} & \multicolumn{1}{c}{\bf Helmet} & \multicolumn{1}{c}{\bf LongBench v2} & \multicolumn{1}{c}{\bf DUDE} \\
  \midrule
  Length (100B) & 84.7 \textcolor{teal}{(+4.1)} & 83.3 \textcolor{teal}{(+6.2)} & 44.0 \textcolor{teal}{(+2.7)} & 70.5 \textcolor{teal}{(+4.1)} & 70.9 \textcolor{teal}{(+3.1)} & 50.2 \textcolor{teal}{(+13.1)} & 42.0 \textcolor{teal}{(+3.0)} & 56.1 \textcolor{teal}{(+3.3)} \\
  Length-Difficulty (100B) & 84.4 \textcolor{teal}{(+3.8)} & 84.6 \textcolor{teal}{(+7.5)} & 42.7 \textcolor{teal}{(+1.3)} & 69.3 \textcolor{teal}{(+2.9)} & 68.0 \textcolor{teal}{(+0.2)} & 51.7 \textcolor{teal}{(+14.6)} & 47.0 \textcolor{teal}{(+8.0)} & 60.1 \textcolor{teal}{(+7.3)} \\
  \midrule
  No Curriculum (10B) & 83.7 \textcolor{teal}{(+3.1)} & 81.2 \textcolor{teal}{(+4.1)} & 43.9 \textcolor{teal}{(+2.5)} & 68.6 \textcolor{teal}{(+2.2)} & 66.1 \textcolor{red}{(-1.7)} & 46.7 \textcolor{teal}{(+9.7)} & 39.0 & 57.6 \textcolor{teal}{(+4.8)} \\
  Length-Difficulty (10B) & 83.4 \textcolor{teal}{(+2.8)} & 81.3 \textcolor{teal}{(+4.2)} & 43.1 \textcolor{teal}{(+1.7)} & 68.1 \textcolor{teal}{(+1.7)} & 66.9 \textcolor{red}{(-0.9)} & 46.0 \textcolor{teal}{(+8.9)} & 41.0 \textcolor{teal}{(+2.0)} & 55.1 \textcolor{teal}{(+2.3)} \\
  Length (10B) & 82.7 \textcolor{teal}{(+2.1)} & 81.3 \textcolor{teal}{(+4.2)} & 41.3 \textcolor{red}{(-0.1)} & 65.0 \textcolor{red}{(-1.4)} & 70.4 \textcolor{teal}{(+2.6)} & 48.6 \textcolor{teal}{(+11.6)} & 41.0 \textcolor{teal}{(+2.0)} & 60.7 \textcolor{teal}{(+7.9)} \\
  \bottomrule
  \end{tabular}
  }%
  \end{center}
  \caption{Curriculum comparison after continued pretraining (CPT), deltas shown from the base model (Mistral Base).}\label{tab:cpt_curricula}
\end{table}

\subsubsection{CPT Qwen3 VL}
\label{sec:cpt_qwen3vl}
We apply CPT to Qwen3 VL for 10B tokens with the same data and find that evaluation scores improve in similar fashion to CPT on Mistral (see Table~\ref{tab:cpt_scale}). Specifically, we see MMLBD-C, MMLB 128K and HELMET scores increase. Given Qwen3 VL's already SOTA performance on MMLongBenchDoc and the difficulty of the benchmark, improvements from the unsupervised CPT tasks are impressive.

\subsubsection{Extended context length analysis}
\label{sec:extended_context_analysis}

Prompted by the success of SFT only compared to CPT + SFT on MMLBD-C, we explore the strength of Mistral on extended context length examples for the two scenarios. Note that Mistral Instruct has a context length of 128K and that the large majority of MMLongBenchDoc examples are under 100 pages which falls within this context and may explain the success of SFT only on our evaluations. We evaluate SFT from CPT and SFT only on a toy benchmark consisting of examples with a set number of pages and a question and ground truth answer generated from a single page from the example. We use a LLM Judge and measure the performance at 150 pages and 300 pages per example. Scored from 1-5, SFT from CPT achieves 3.75 and 3.45 respectively and SFT only achieves 3.53 and 3.52 respectively. In this dataset, there are only 2400 examples with more than 150 pages, so we are surprised to find that SFT adapts very quickly to the extended context length. There is a non-negligible difference in performance between CPT and SFT only at 150 pages, suggesting CPT improves LC performance mainly at context lengths far below the maximum context length seen in CPT.

\begin{figure}[t]
\begin{center}
\includegraphics[width=0.6\linewidth]{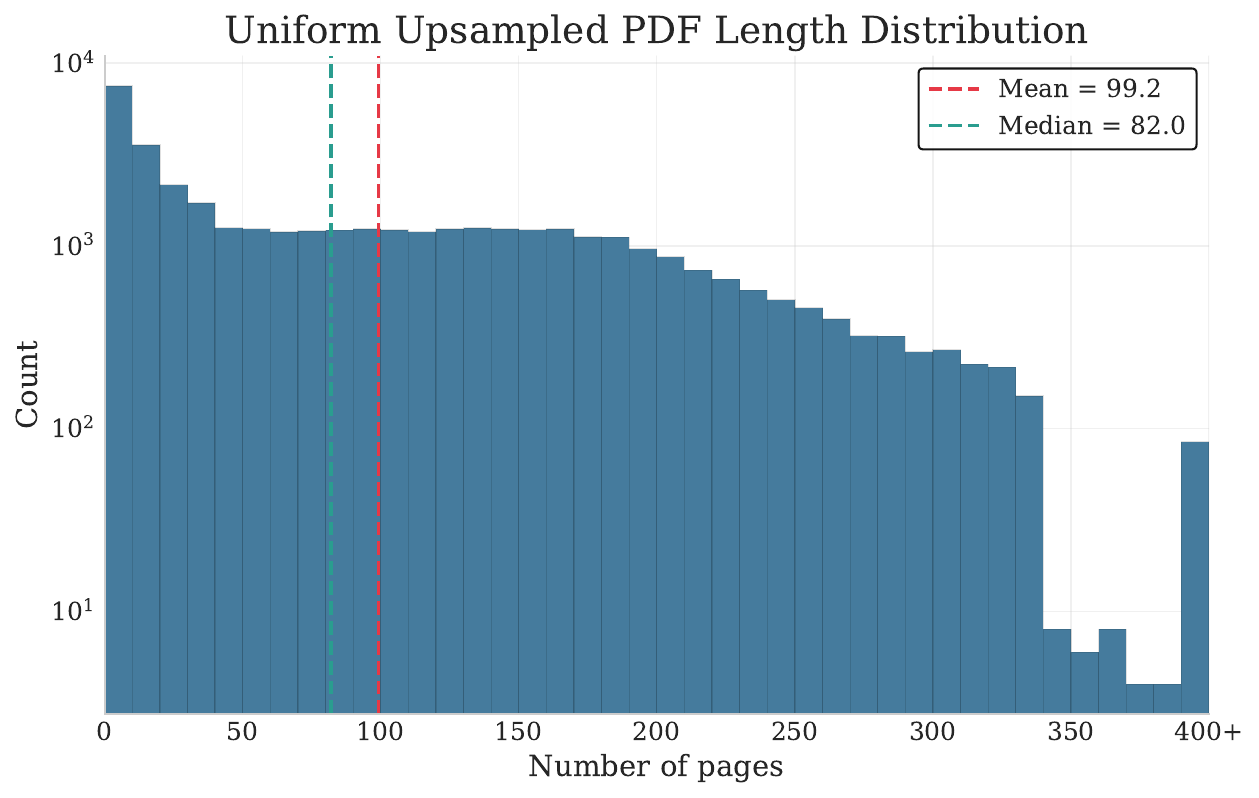}
\end{center}
\caption{PDF page-length distribution after upsampling long documents for CPT (compared to the natural scraped-corpus distribution in Figure~\ref{fig:scraped_corpus_overview}).}
\label{fig:upsampled_long_length_dist}
\end{figure}

\paragraph{Upsampling long documents in CPT.}
We attempted CPT with upsampled long documents, with the distribution shifted close to uniform (see \ref{fig:upsampled_long_length_dist}) and found that this degraded performance. We attribute this to the finding that training on the short stage only yields better performance, though we did not test targeting different distributions or domain normalized upsampling. Results shown in Table~\ref{tab:cpt_upsampled_long_pdfs}.

\begin{table}[t]
  \begin{center}
  \resizebox{\textwidth}{!}{%
  \begin{tabular}{lccccccccc}
  \toprule
  \multicolumn{1}{l}{\bf Checkpoint} & \multicolumn{1}{c}{\bf VA} & \multicolumn{1}{c}{\bf LCA} & \multicolumn{1}{c}{\bf MMLBD-C} & \multicolumn{1}{c}{\bf MMLB 128K} & \multicolumn{1}{c}{\bf MMLB 32K} & \multicolumn{1}{c}{\bf SlideVQA} & \multicolumn{1}{c}{\bf Helmet} & \multicolumn{1}{c}{\bf LongBench v2} & \multicolumn{1}{c}{\bf DUDE} \\
  \midrule
  CPT & 83.4 & 81.3 & 43.1 & 68.1 & 75.0 & 66.9 & 46.0 & 41.0 & 55.1 \\
  CPT Upsampled Long PDFs & 82.1 \textcolor{red}{(-1.3)} & 81.3 & 41.7 \textcolor{red}{(-1.4)} & 70.3 \textcolor{teal}{(+2.2)} & 73.4 \textcolor{red}{(-1.6)} & 66.3 \textcolor{red}{(-0.7)} & 45.8 \textcolor{red}{(-0.2)} & 45.0 \textcolor{teal}{(+4.0)} & 54.6 \textcolor{red}{(-0.5)} \\
  \bottomrule
  \end{tabular}
  }%
  \end{center}
  \caption{CPT with upsampled long documents vs CPT with natural distribution.}\label{tab:cpt_upsampled_long_pdfs}
\end{table}

\subsection{Supervised finetuning (SFT) experiments}
\label{sec:sft_experiments}
\subsubsection{Question generation details}
\label{sec:question_generation_details}
We develop a novel question generation pipeline targeting multi-page questions, i.e. questions that require evidence from multiple pages to answer correctly, to teach the model to aggregate information across the document, and combine this with a simple pipeline for single-page question generation which is cheaper to scale and targets retrieval and general QA capabilities in the model.

\label{pa:magpie}
As a baseline, we employ Magpie for its simplicity and effectiveness. Magpie simply provides the page to the VLM and generates a completion which is usually a simulated user question. For both question pipelines, we generate answers using Qwen3 VL 235B A22B given the full context. Our pipelines yield minor VA and LCA improvements but degrade long-document performance on MMLBD-C and DUDE. See Table~\ref{tab:magpie_vs_sp_mp} for details.

\paragraph{Single-page questions.}
We prompt the model with a randomly selected page to generate a varying number of questions, including a randomly selected question archetype prompt for each question, e.g. ``ask a difficult question that has a short, verifiable (not open-ended or debatable, has a single correct answer) answer (a number, string, list, dictionary, yes/no, etc.) and ask for the model to reason before answering'', then select one of those questions to keep. Varying the number of questions ensures the final question set avoids mode averaging, i.e. asking the expected question for the page. An example needing an answer can be constructed from a subsection of adjacent pages within the document, the entire document or hard negative pages.

\paragraph{Multi-page questions.}
The multi-page question pipeline extends this by providing a set of pages that can be drawn from an adjacent range within a document, random pages from across the document or hard negatives along with a prompt to generate questions that require evidence from multiple pages to answer. To filter for questions that fulfill this requirement, we use a smaller VLM, Qwen2.5 VL 7B \citet{qwen25vl} or Qwen3 VL 32B, to answer the question given each page individually. A judge determines whether the question has been fully and correctly answered and we keep only questions which were not correctly answered with any single page. The remaining questions are more likely to require aggregating information from multiple pages.

\paragraph{Single-page vs multi-page questions.}
We compare the performance of SFT with single-page questions only vs multi-page questions only. As shown in Table \ref{tab:single_page_vs_multi_page_questions}, we find that multi-page questions perform worse on MMLongBench, indicating the score focuses more on retrieval capabilities than cross-page reasoning. On MMLBD-C, where there are explicit single-page and multi-page question types, scores are similar. The two types of questions complement each other for more robust performance.

\begin{table}[t]
  \begin{center}
  \resizebox{\textwidth}{!}{%
  \begin{tabular}{lcccccccc}
  \toprule
  \multicolumn{1}{l}{\bf Checkpoint} & \multicolumn{1}{c}{\bf VA} & \multicolumn{1}{c}{\bf LCA} & \multicolumn{1}{c}{\bf MMLBD-C} & \multicolumn{1}{c}{\bf MMLB 128K} & \multicolumn{1}{c}{\bf SlideVQA} & \multicolumn{1}{c}{\bf Helmet} & \multicolumn{1}{c}{\bf LongBench v2} & \multicolumn{1}{c}{\bf DUDE} \\
  \midrule
  Single-Page & 82.0 \textcolor{teal}{(+1.7)} & 79.1 \textcolor{teal}{(+2.7)} & 43.0 \textcolor{teal}{(+0.2)} & 68.3 \textcolor{teal}{(+6.7)} & 62.0 \textcolor{red}{(-5.1)} & 40.0 \textcolor{red}{(-0.6)} & 41.0 \textcolor{teal}{(+6.0)} & 53.0 \textcolor{red}{(-2.7)} \\
  Multi-Page & 80.3 & 76.4 & 42.8 & 61.5 & 67.2 & 40.6 & 35.0 & 55.7 \\
  \bottomrule
  \end{tabular}
  }%
  \end{center}
  \caption{Single-page vs multi-page questions in SFT.}\label{tab:single_page_vs_multi_page_questions}
\end{table}

\begin{table}[t]
  \begin{center}
  \resizebox{\textwidth}{!}{%
  \begin{tabular}{lcccccccc}
  \toprule
  \multicolumn{1}{l}{\bf Checkpoint} & \multicolumn{1}{c}{\bf VA} & \multicolumn{1}{c}{\bf LCA} & \multicolumn{1}{c}{\bf MMLBD-C} & \multicolumn{1}{c}{\bf MMLB 128K} & \multicolumn{1}{c}{\bf SlideVQA} & \multicolumn{1}{c}{\bf Helmet} & \multicolumn{1}{c}{\bf LongBench v2} & \multicolumn{1}{c}{\bf DUDE} \\
  \midrule
  SP + MP & 84.1 \textcolor{teal}{(+0.3)} & 82.8 \textcolor{teal}{(+0.8)} & 45.8 \textcolor{red}{(-1.6)} & 69.0 \textcolor{teal}{(+2.8)} & 70.5 \textcolor{teal}{(+0.7)} & 58.0 \textcolor{teal}{(+3.2)} & 36.0 & 52.2 \textcolor{red}{(-1.4)} \\
  Magpie & 83.8 & 82.0 & 47.5 & 66.2 & 69.8 & 54.8 & 36.0 & 53.6 \\
  \bottomrule
  \end{tabular}
  }%
  \end{center}
  \caption{Comparison of question generation pipelines: magpie vs our single-page and multi-page questions.}\label{tab:magpie_vs_sp_mp}
\end{table}

\subsubsection{Alternative synthetic data pipelines}
\label{sec:appendix_alternative_synthetic_data_pipelines}
We briefly describe some of the alternative answer generation pipelines we developed, which are involved in some of the earlier checkpoints we train.

\begin{itemize}
  \item Distractors short: We take hard negatives from outside the first 32, intended to be similar but not enough to have information content that should be included in the answer and form examples up to 5 pages in length. We use a VLM to answer a SP question based only on the page used to generate the question.
  \item Adjacent short: We take a subset of adjacent pages from a document between 2-5 pages in length and use a VLM to generate the answer given the full context.
  \item HN short: We take hard negatives from within the first 32, construct examples up to 5 pages in length and use a VLM to generate the answer given the full context.
  \item Multi-turn: We simulate a multi-turn conversation with SP and MP question prompts (excluding the MP question verification step), also prompting the model to either probe deeper or ask a new question, and generate answers with the recursive pipeline. We also add examples constructed by simply concatenating single-turn examples from other pipelines.
  \item Unanswerable: We prompt the model to generate trick questions. We found this harmful to MMLBD-C performance which was the target and upon inspection, noticed that these questions appear naturally in our recursive pipeline so we excluded it from future runs.
  \item Quality filter: We experiment with a quality filter pipeline and find only minor improvements in VA. The quality filter adapts the recursive pipeline to the task of checking for inconsistencies between the answer and the document. It breaks an answer down into a list of assertions and collects evidence from each page relevant to the assertions. It then uses a VLM to check if the most relevant pages and the extracted evidence are enough to support all assertions. We note this pipeline not only filters data, but provides a new task for the model to learn from, i.e. the final check is a visual/text LC task. We did not experiment with the impact of this task.
\end{itemize}

\subsubsection{Additional SFT experiments}
\label{sec:additional_sft_experiments}

\begin{table}[t]
  \begin{center}
  \begin{tabular}{lccccc}
  \toprule
  \textbf{Dataset} & \textbf{Unit} & \textbf{Mean} & \textbf{Median} & \textbf{Max} & \textbf{N Examples} \\
  \midrule
  Ours: Short Stage & images & 21.1 & 9 & 104 & 52,433 \\
  Ours: Long Stage & images & 145.3 & 156 & 336 & 22,076 \\
  \midrule
  ProLong: 64K & tokens & 1,350 & 533 & 64K & 83.8M \\
  ProLong: 512K & tokens & 1,262 & 484 & 512K & 60.4M \\
  \bottomrule
  \end{tabular}
  \end{center}
  \caption{Comparison of training data length distributions. ProLong's 512K stage is heavily short-skewed (median of 484 tokens despite 512K max), while our long stage contains genuinely long examples (median of 156 images). This explains why ProLong benefits from their ``longer'' training while we observe degradation.}\label{tab:prolong_comparison}
\end{table}

\paragraph{Base model}
\label{pa:base_model}
GRAPE \citet{grape} is a recent work that shows that SFT data that matches the base model's distribution more closely is more effective. Extrapolating from this and from common practice, we hypothesize that the instruct model will perform better than the base model for our training. However, in this work we make extensive use of model merging and we lack guidance on the expected performance of applying SFT to the CPT merged model vs the instruct model followed by merging. Thus, we compare the performance of SFT: from the base model vs the instruct model vs the merged CPT model. In each case, we apply the respective instruct/CPT vectors to get a final checkpoint that is a combination of the default instruct tuned version, the CPT vector and the new LC SFT vector. VA scores in Table \ref{tab:base_model} show that SFT from the instruct model is indeed stronger than SFT from the base model and additionally that SFT from the merged CPT checkpoint significantly outperforms SFT from the instruct model followed by adding the CPT vector.

\begin{table}[t]
  \begin{center}
  \resizebox{\textwidth}{!}{%
  \begin{tabular}{lcccccccc}
  \toprule
  \multicolumn{1}{l}{\bf Checkpoint} & \multicolumn{1}{c}{\bf VA} & \multicolumn{1}{c}{\bf LCA} & \multicolumn{1}{c}{\bf MMLBD-C} & \multicolumn{1}{c}{\bf MMLB 128K} & \multicolumn{1}{c}{\bf SlideVQA} & \multicolumn{1}{c}{\bf Helmet} & \multicolumn{1}{c}{\bf LongBench v2} & \multicolumn{1}{c}{\bf DUDE} \\
  \midrule
  Instruct + CPT & 84.6 \textcolor{teal}{(+4.7)} & 82.7 \textcolor{teal}{(+5.1)} & 46.0 \textcolor{teal}{(+3.0)} & 64.5 \textcolor{teal}{(+5.6)} & 70.0 \textcolor{teal}{(+3.0)} & 49.8 \textcolor{teal}{(+0.4)} & 40.0 \textcolor{teal}{(+6.0)} & 55.4 \textcolor{teal}{(+1.0)} \\
  Instruct & 83.3 \textcolor{teal}{(+3.4)} & 82.5 \textcolor{teal}{(+4.9)} & 45.0 \textcolor{teal}{(+2.1)} & 65.1 \textcolor{teal}{(+6.2)} & 67.6 \textcolor{teal}{(+0.5)} & 54.0 \textcolor{teal}{(+4.6)} & 40.0 \textcolor{teal}{(+6.0)} & 56.2 \textcolor{teal}{(+1.8)} \\
  Base & 79.9 & 77.6 & 42.9 & 58.9 & 67.1 & 49.5 & 34.0 & 54.4 \\
  \bottomrule
  \end{tabular}
  }%
  \end{center}
  \caption{SFT from the base model vs the instruct model vs the merged CPT model.}\label{tab:base_model}
\end{table}

\paragraph{Prompting for answer generation.}
\label{pa:prompting_for_answer_generation}
One of the details of the plain distillation pipeline is the lack of prompting for the model aside from the default system prompt. Given the strong results, we minimize prompting for answers across all our pipelines.

\paragraph{SFT Scale}
\label{pa:sft_scale}
We ablate the number of examples in SFT, comparing runs with 10K and 50K examples on Mistral with the same data and settings. As shown in Table \ref{tab:sft_scale}, 50K examples is significantly better than 10K examples, achieving a 2.2 point improvement in VA and a 1.5 point improvement in LCA. For visual long-document VQA performance as measured on MMLBD-C, 10K examples is enough for maximum performance.

\begin{table}[t]
  \begin{center}
  \resizebox{\textwidth}{!}{%
  \begin{tabular}{lcccccccc}
  \toprule
  \multicolumn{1}{l}{\bf Checkpoint} & \multicolumn{1}{c}{\bf VA} & \multicolumn{1}{c}{\bf LCA} & \multicolumn{1}{c}{\bf MMLBD-C} & \multicolumn{1}{c}{\bf MMLB 128K} & \multicolumn{1}{c}{\bf SlideVQA} & \multicolumn{1}{c}{\bf Helmet} & \multicolumn{1}{c}{\bf LongBench v2} & \multicolumn{1}{c}{\bf DUDE} \\
  \midrule
  SFT (50K) & 84.1 \textcolor{teal}{(+2.1)} & 82.8 \textcolor{teal}{(+1.5)} & 45.8 \textcolor{teal}{(+0.6)} & 69.0 \textcolor{teal}{(+10.4)} & 70.5 \textcolor{red}{(-1.5)} & 58.0 \textcolor{teal}{(+3.6)} & 36.0 \textcolor{red}{(-3.0)} & 52.2 \textcolor{red}{(-0.9)} \\
  SFT (10K) & 82.0 & 81.3 & 45.2 & 58.5 & 72.0 & 54.4 & 39.0 & 53.1 \\
  \bottomrule
  \end{tabular}
  }%
  \end{center}
  \caption{SFT scale ablation.}\label{tab:sft_scale}
\end{table}

\paragraph{Impact of training in two stages.}
We split training into two stages for efficiency, first training on up to 104 pages and then training on up to 336 pages. To measure the impact of this decision, we compare training on the same data, split in two stages vs all in one stage. For single stage training, the data order is fully shuffled. As shown in Table \ref{tab:two_stage_training}, we find that training in a single stage is most effective, yielding a 1.3 point improvement in VA and a 2.2 point improvement in LCA. However, given the additional overhead of increased parallelism at higher context lengths, we train in two stages.

\begin{table}[t]
  \begin{center}
  \resizebox{\textwidth}{!}{%
  \begin{tabular}{lcccccccccc}
  \toprule
  \multicolumn{1}{l}{\bf Checkpoint} & \multicolumn{1}{c}{\bf VA} & \multicolumn{1}{c}{\bf LCA} & \multicolumn{1}{c}{\bf MMLBD} & \multicolumn{1}{c}{\bf MMLBD-C} & \multicolumn{1}{c}{\bf MMLB 128K} & \multicolumn{1}{c}{\bf MMLB 32K} & \multicolumn{1}{c}{\bf SlideVQA} & \multicolumn{1}{c}{\bf Helmet} & \multicolumn{1}{c}{\bf LongBench v2} & \multicolumn{1}{c}{\bf DUDE} \\
  \midrule
  Single Stage & 82.8 \textcolor{teal}{(+1.3)} & 82.5 \textcolor{teal}{(+2.2)} & 43.2 \textcolor{red}{(-0.1)} & 66.0 \textcolor{teal}{(+1.3)} & 73.6 \textcolor{teal}{(+1.3)} & 68.1 \textcolor{teal}{(+2.2)} & 57.4 \textcolor{teal}{(+6.9)} & 39.0 & 56.5 \textcolor{teal}{(+2.1)} \\
  Two Stages & 81.5 & 80.2 & 43.3 & 64.7 & 72.2 & 65.9 & 50.6 & 39.0 & 54.4 \\
  \bottomrule
  \end{tabular}
  }%
  \end{center}
  \caption{Impact of training in two stages.}\label{tab:two_stage_training}
\end{table}

\paragraph{Impact of external SFT data.}
We compare the performance impact of training on external SFT data with two ablations on Mistral: training with 25K examples of our synthetic data vs adding 25K examples of external SFT data (from Smoltalk2 \citet{smoltalk2}, Luth \citet{luthsft}, DocFinQA \citet{docfinqa} and ChartQA \citet{chartqa}) vs adding 400K examples of external SFT data. As shown in Table \ref{tab:external_sft_data}, we notice that a small amount of external SFT data is harmful to VA and LCA, while a large amount of external SFT data is slightly beneficial. We found the same results for 25K examples of external SFT data in the self-improving setting.

\begin{table}[t]
  \begin{center}
  \resizebox{\textwidth}{!}{%
  \begin{tabular}{lccccccccc}
  \toprule
  \multicolumn{1}{l}{\bf Checkpoint} & \multicolumn{1}{c}{\bf VA} & \multicolumn{1}{c}{\bf LCA} & \multicolumn{1}{c}{\bf MMLBD-C} & \multicolumn{1}{c}{\bf MMLB 128K} & \multicolumn{1}{c}{\bf MMLB 32K} & \multicolumn{1}{c}{\bf SlideVQA} & \multicolumn{1}{c}{\bf Helmet} & \multicolumn{1}{c}{\bf LongBench v2} & \multicolumn{1}{c}{\bf DUDE} \\
  \midrule
  400K External SFT & 82.9 \textcolor{teal}{(+0.2)} & 81.5 \textcolor{teal}{(+0.1)} & 44.6 \textcolor{teal}{(+0.9)} & 70.4 \textcolor{teal}{(+5.1)} & 73.3 \textcolor{red}{(-0.7)} & 61.7 \textcolor{red}{(-7.3)} & 54.4 \textcolor{teal}{(+3.7)} & 37.0 \textcolor{red}{(-3.0)} & 54.9 \textcolor{teal}{(+0.1)} \\
  No External SFT & 82.7 & 81.4 & 43.7 & 65.3 & 74.1 & 69.1 & 50.7 & 40.0 & 54.8 \\
  25K External SFT & 81.5 \textcolor{red}{(-1.2)} & 80.2 \textcolor{red}{(-1.2)} & 43.3 \textcolor{red}{(-0.4)} & 64.7 \textcolor{red}{(-0.6)} & 72.2 \textcolor{red}{(-1.8)} & 65.9 \textcolor{red}{(-3.2)} & 50.6 \textcolor{red}{(-0.2)} & 39.0 \textcolor{red}{(-1.0)} & 54.4 \textcolor{red}{(-0.4)} \\
  \bottomrule
  \end{tabular}
  }%
  \end{center}
  \caption{Impact of external SFT data.}\label{tab:external_sft_data}
\end{table}

\paragraph{PoSE.}
We tested PoSE \citet{pose} with a target context length of 1M tokens and found that it degrades VA by -2.0 points. We did not attempt further investigation or use PoSE in the rest of our experiments.

\paragraph{Hard negatives.}
\label{sec:hard_negatives}
We ablate the impact of hard negative examples by comparing SFT on hard negative examples vs documents only. As shown in Table \ref{tab:hard_negatives}, hard negatives provide a noticeable improvements in VA and LCA. However, we note from earlier that training on only the short stage (up to 104 page examples) outperforms training on both stages (up to 336 pages) and that the hard negative examples we construct are all less than 104 pages, while the documents include longer examples. Generally, we recommend hard negative examples for diverse inputs, the ability to expand the number of examples that can be constructed from a given set of pages and tentative improvements in performance.

\begin{table}[t]
  \begin{center}
  \resizebox{\textwidth}{!}{%
  \begin{tabular}{lcccccccc}
  \toprule
  \multicolumn{1}{l}{\bf Checkpoint} & \multicolumn{1}{c}{\bf VA} & \multicolumn{1}{c}{\bf LCA} & \multicolumn{1}{c}{\bf MMLBD-C} & \multicolumn{1}{c}{\bf MMLB 128K} & \multicolumn{1}{c}{\bf SlideVQA} & \multicolumn{1}{c}{\bf Helmet} & \multicolumn{1}{c}{\bf LongBench v2} & \multicolumn{1}{c}{\bf DUDE} \\
  \midrule
  Hard Negatives & 83.6 \textcolor{teal}{(+1.6)} & 82.5 \textcolor{teal}{(+2.6)} & 44.0 \textcolor{teal}{(+1.6)} & 64.5 \textcolor{red}{(-6.0)} & 68.5 \textcolor{teal}{(+0.5)} & 47.4 \textcolor{teal}{(+2.1)} & 44.0 \textcolor{teal}{(+4.0)} & 56.7 \textcolor{teal}{(+1.0)} \\
  Documents Only & 82.0 & 79.9 & 42.4 & 70.6 & 68.1 & 45.3 & 40.0 & 55.7 \\
  \bottomrule
  \end{tabular}
  }%
  \end{center}
  \caption{Hard negatives vs documents in SFT.}\label{tab:hard_negatives}
\end{table}

\subsection{LongPO}
\subsubsection{Recursive vs plain distillation}
\label{sec:longpo_recursive_vs_plain_distill}
We compare the performance of the recursive pipeline vs plain distillation in the LongPO setting, using the same data as in \nameref{sec:answer_generation}. As shown in Table \ref{tab:longpo_recursive_vs_plain_distill}, the results are essentially identical. We previously noted the limitations of answers generated from 'question pages' only and while we find a slight improvement in VA in SFT, LongPO results indicate that both methods are equally effective.

\begin{table}[t]
  \begin{center}
  \resizebox{\textwidth}{!}{%
  \begin{tabular}{lcccccccc}
  \toprule
  \multicolumn{1}{l}{\bf Checkpoint} & \multicolumn{1}{c}{\bf VA} & \multicolumn{1}{c}{\bf LCA} & \multicolumn{1}{c}{\bf MMLBD-C} & \multicolumn{1}{c}{\bf MMLB 128K} & \multicolumn{1}{c}{\bf SlideVQA} & \multicolumn{1}{c}{\bf Helmet} & \multicolumn{1}{c}{\bf LongBench v2} & \multicolumn{1}{c}{\bf DUDE} \\
  \midrule
  Plain Distillation & 92.4 & 91.5 & 54.0 & 71.8 & 74.8 & 63.4 & 42.0 & 54.1 \\
  Recursive & 92.2 \textcolor{red}{(-0.2)} & 90.9 \textcolor{red}{(-0.6)} & 54.3 \textcolor{teal}{(+0.3)} & 73.2 \textcolor{teal}{(+1.4)} & 74.7 \textcolor{red}{(-0.1)} & 62.4 \textcolor{red}{(-1.0)} & 41.0 \textcolor{red}{(-1.0)} & 54.1 \\
  \bottomrule
  \end{tabular}
  }%
  \end{center}
  \caption{Recursive vs Plain Distillation in LongPO setting.}\label{tab:longpo_recursive_vs_plain_distill}
\end{table}

\subsection{Extended suggestions}
\label{sec:extended_suggestions}
Based on the large number of ablations we performed and the key performance factors we identify, we summarize the highest-signal findings and provide the following condensed recommendations for training long-context visual document models (a more comprehensive list can be found in Appendix~\ref{sec:sft_experiments}):
\begin{itemize}
  \item Train on all CPT tasks, including LC text data, for best performance, or exclude counting to keep data scalable with minimal loss of performance. See \hyperref[sec:cpt_tasks]{Impact of each CPT task}.
  \item Train from the Instruct when not performing CPT, otherwise the Instruct + CPT checkpoint. See \hyperref[pa:base_model]{SFT base model}.
  \item In LongPO, it is sufficient to use the strongest teacher available (including the model itself), the recursive answer generation pipeline is not necessary. See \hyperref[tab:longpo_recursive_vs_plain_distill]{LongPO recursive vs plain distillation}.
\end{itemize}

\subsection{External SFT data composition}
\label{sec:external_sft_composition}

Tables~\ref{tab:luth_composition} and~\ref{tab:smoltalk2_composition} detail the normalized composition of the external SFT data used in our experiments. We draw samples according to the distributions in the table.

\begin{table}[h]
  \centering
  \begin{tabular}{lc}
  \toprule
  \textbf{Source} & \textbf{Proportion (\%)} \\
  \midrule
  Scholar & 30.0 \\
  Smoltalk2 & 30.0 \\
  Aya Dataset & 10.0 \\
  Tulu3 Persona Math & 10.0 \\
  Tulu3 Persona Instruct & 10.0 \\
  OpenHermes & 10.0 \\
  \bottomrule
  \end{tabular}
\caption{Composition of the Luth \citet{luthsft} SFT data mixture.}
\label{tab:luth_composition}
\end{table}

\begin{table}[h]
  \centering
  \begin{tabular}{lc}
  \toprule
  \textbf{Source} & \textbf{Proportion (\%)} \\
  \midrule
  LongAlign 64K & 18.0 \\
  Mixture of Thoughts (Science) & 18.0 \\
  OpenThoughts3 1.2M & 18.0 \\
  TableGPT & 18.0 \\
  Tulu3 SFT Personas Instruction Following & 9.0 \\
  Smoltalk Multilingual (8 languages) & 3.6 \\
  Smoltalk Smol Magpie Ultra & 3.6 \\
  Smoltalk Smol Summarize & 3.6 \\
  Multifaceted Collection & 3.6 \\
  OpenHermes 2.5 & 1.8 \\
  EverythingLM-data-V3 & 1.8 \\
  Smoltalk Everyday Conversations & 0.9 \\
  \bottomrule
  \end{tabular}
  \caption{Composition of the Smoltalk2 \citet{smoltalk2} SFT data mixture.}
  \label{tab:smoltalk2_composition}
\end{table}

\end{document}